%% file: main.tex
\documentclass{article} %
\usepackage{iclr2027_conference,times}
\iclrfinalcopy

\input{math_commands.tex}

\usepackage{hyperref}
\usepackage{url}
\usepackage{booktabs} %
\usepackage{graphicx}
\usepackage{array}
\usepackage{wrapfig}
\usepackage{subcaption} %
\usepackage{enumitem}
\title{La-JEPA: Compact Robot Policies with Latent World Modeling}

\author{Mateo Cl\'{e}mente$^{1,\dagger}$, George Z.~Xue$^{2,*}$, Amin Abyaneh$^{3,*}$, Liheng Ma$^{4}$, Rui Heng Yang$^{1}$, \\
\textbf{Tongtong Cao$^{1}$, Zhanguang Zhang$^{1}$, Yingxue Zhang$^{1}$} \\
\textnormal{$^{1}$Huawei Technologies Ltd. \quad $^{2}$University of Toronto \quad $^{3}$McGill University} \\
\textnormal{$^{4}$Mila -- Quebec Artificial Intelligence Institute}
}

\begin{document}

\maketitle
{\renewcommand{\thefootnote}{$\dagger$}\footnotetext{Corresponding author: \texttt{mateo.clem@gmail.com}.}}
{\renewcommand{\thefootnote}{$*$}\footnotetext{As interns for Huawei.}}
{\renewcommand{\thefootnote}{}\footnotetext{Code: \url{https://github.com/mt-clemente/lajepa}}}
\lhead{Preprint}

\begin{abstract}
World-action models commonly adapt pretrained video generators, and vision-language-action models build on pretrained vision-language backbones; both tie policy architecture and inference cost to large transformers designed for generation rather than control. While the knowledge accumulated through large-scale pretraining is certainly useful, it remains unclear how much generation-oriented architectural choices affect control performance, and whether the trade-off required to access this pretraining is worth it. While recent world-action models mostly scale the video backbone and its data, we argue that the choice of latent space is just as critical, and that the policy should be trained natively in that space. In this work, we introduce \textbf{Latent Action JEPA (La-JEPA)}, a compact world-action model whose policy transformer is trained from scratch in the feature space of a pretrained V-JEPA~2.1 encoder, so that its architecture and capacity can be chosen for control. Following Fast-WAM, La-JEPA learns to predict future features jointly with actions during training and generates only actions at inference. \textbf{Among policies without embodied pretraining, La-JEPA achieves state-of-the-art performance on RoboTwin~2.0, in clean and randomized scenes when trained on clean demonstrations and on average when trained on the full training set, and on LIBERO-Plus with corrected instructions, while using far fewer parameters than the strongest competing policies.} With full training, its success also increases monotonically with model size, showing that the approach scales. Its S and B policies run two to five times faster than Fast-WAM and Faster-WAM while outperforming both on LIBERO-Plus and RoboTwin~2.0. La-JEPA also outperforms the closest JEPA-based methods, LeapBot-WA and JEPA-WAM, on LIBERO-Plus and RoboTwin~2.0 without their pretrained policy backbones. On four real-robot tasks, including contact-heavy pushing and long-horizon multistep manipulation, its average success is comparable to $\pi_{0.5}$, outperforming non-embodied-pretrained baselines. Through a design study of the choices behind La-JEPA, we show that JEPA features outperform video VAE latents, most clearly out of distribution, and that training-time future prediction choices significantly impact the policy even when future prediction is skipped at inference. These results establish a compact, efficient foundation for JEPA-based world-action modeling: without a generative pretrained policy backbone, La-JEPA is better and faster than far larger policies built on video and language generation models. \textbf{Tailoring the policy's design, including its latent space, to control can thus matter more than the knowledge transferred from a heavily pretrained generative backbone}, suggesting that robot foundation models may benefit from architectures designed specifically for control rather than inherited from generation.
\end{abstract}

\section{Introduction}
\label{sec:introduction}

World-action models increasingly adapt pretrained video-generation transformers to robot control, inheriting useful visual and temporal priors along with the representation space, tokenization, and architecture of a model built for video synthesis~\citep{fastwam,fasterwam,cosmos_policy}. Vision-language-action policies take a similar route using language-generation transformers aligned with visual inputs~\citep{openvla,pi0,pi05}. The success of these two types of architectures raises two practical questions: must robot policies necessarily inherit the architecture of a generative backbone to be competitive, and can a compact world-action transformer trained from scratch actually reach the performance of these policies?

\begin{figure}[t]
\centering
\includegraphics[width=\linewidth]{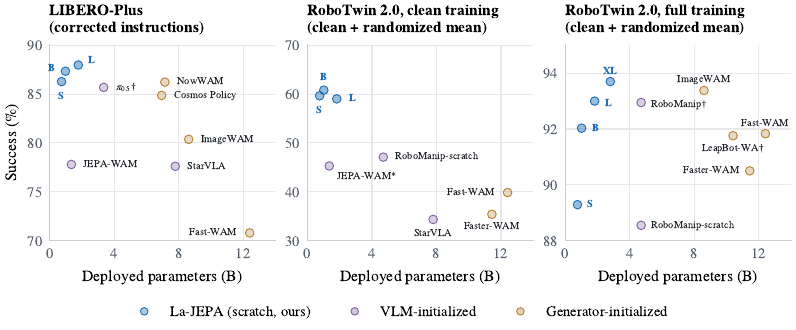}
\caption{Success against deployed parameters, including frozen encoders, on LIBERO-Plus with corrected instructions (Section~\ref{sec:libero-plus-text}) and on RoboTwin~2.0, where the panels report the equally weighted mean of clean and randomized success of policies trained on clean demonstrations only (center) or on the full clean and randomized training set (right). S, B, L, and XL denote the La-JEPA sizes. The LIBERO-Plus panel shows the policies without embodied pretraining that we re-evaluated with corrected instructions, and the embodied-pretrained $\pi_{0.5}$; the full-training panel adds Qwen-RobotManip (RoboManip) and its scratch variant~\citep{qwen_robotmanip} and LeapBot-WA~\citep{leapbot}, as reported by their authors. $\dagger$: embodied pretraining. Table~\ref{tab:main-comparison} lists all scores, including those with the released instructions. $*$: success estimated from JEPA-WAM's 20-task results (Appendix~\ref{app:protocols}).}
\label{fig:main-efficiency}
\end{figure}

Most world-action models (WAMs) are VAE-based because they build on pretrained video generators, yet better-suited latent spaces are available. Video VAEs retain accurate information but spend much of their capacity on pixel-level fidelity that control does not require. JEPA representations, in contrast, are developed and evaluated for downstream tasks such as video understanding, depth estimation, and segmentation, which are closer to the needs of embodied agents~\citep{vjepa2,vjepa21}. Previous work aligns VLMs or video-generation models to the JEPA space~\citep{jepawam,leapbot} or learns a JEPA-style latent space jointly with the policy~\citep{actjepa}; to our knowledge, no prior work trains a compact world-action policy from scratch in a pretrained JEPA feature space.

We introduce \textbf{Latent Action JEPA (La-JEPA)}, a world-action model that learns future-feature prediction and action generation directly in the dense spatial and temporal feature space of a pretrained V-JEPA~2.1 video encoder~\citep{vjepa21}. Its World DiT and Action DiT are initialized from scratch, allowing their architecture and capacity to be chosen independently of the visual encoder. Following Fast-WAM and Faster-WAM~\citep{fastwam,fasterwam}, actions attend to observed context but not to future tokens, so predictive supervision can shape the policy during training without the cost of generating the future at inference time.

Among policies without embodied pretraining, La-JEPA achieves the strongest clean and randomized RoboTwin~2.0 results with clean-only training, the strongest average with full training, where success increases monotonically from S to XL, and the strongest LIBERO-Plus results with corrected instructions (Section~\ref{sec:libero-plus-text}), with about four times fewer deployed parameters than the next-best policies, while running up to 5$\times$ faster than Fast-WAM, whose design we follow (Figure~\ref{fig:main-efficiency}, Tables~\ref{tab:main-comparison} and~\ref{tab:robotwin-comparison}). With the released instructions, it is second only to NowWAM, which is insensitive to the corrupted text (Table~\ref{tab:libero-plus-text}). It also outperforms the closest JEPA-based methods, LeapBot-WA and JEPA-WAM, on LIBERO-Plus and RoboTwin~2.0 (Table~\ref{tab:jepa-comparison}). LeapBot-WA aligns the Fast-WAM architecture with the same V-JEPA~2.1 space and still trails La-JEPA, suggesting that training the policy natively in the adapted space matters, not only the space itself. On real-robot tasks ranging from cube placement to contact-rich pushing and long-horizon manipulation, La-JEPA trained from scratch on the target demonstrations achieves the highest average success among the compared policies, comparable to the embodied-pretrained $\pi_{0.5}$ (Table~\ref{tab:real-robot}).

To understand which design choices enable this performance, we study the impact of the representation space, different targets for predictive supervision, and robustness mechanisms. We adopt a \textit{compact first} approach, making the architecture viable even for a policy with 0.1B trainable parameters, before increasing policy capacity. Comparisons of JEPA features and VAE latents span both benchmarks and multiple model configurations, showing that JEPA features generally outperform VAE latents, with the clearest gains for out of distribution evaluations (Section~\ref{sec:predictive-design}, Appendix~\ref{app:ablations}). Action-only controls separately confirm the benefit of WAM-style future prediction, and studies of temporal encoding, dropout, augmentation, and encoder adaptation complete the recipe. As a secondary contribution, we show that the released LIBERO-Plus evaluator appends perturbation parameters to most instructions, so that most variants apply two perturbations at once. This corruption affects policies to very different degrees, from no change to 19 points, and obscures the robustness study that the benchmark is designed for; we therefore re-evaluate the openly available baselines with corrected instructions under a single ``clean'' protocol that isolates each perturbation as intended (Section~\ref{sec:libero-plus-text}).

\section{Related Work}
\label{sec:related-work}

\paragraph{Vision-language-action policies.} Vision-language-action (VLA) policies extend pretrained vision-language models with action outputs, predicting discretized action tokens~\citep{openvla} or attaching a flow-matching action expert~\citep{pi0}, and broaden their training data to generalize further~\citep{pi05,qwen_robotmanip}. All inherit the size and inference cost of a multi-billion-parameter vision-language backbone.

\paragraph{World-action models.} World-action models (WAMs) instead build on pretrained video or image generators whose representations drive action generation~\citep{cosmos_policy,imagewam,nowwam}, reaching the strongest previously reported LIBERO-Plus results among policies without embodied pretraining. Because generating future frames slows inference, recent WAMs predict the future only during training~\citep{fastwam} and reduce the action expert to a single layer that reads keys and values cached from all layers of the video backbone~\citep{fasterwam}. All retain the generator backbone in the deployed policy.

\paragraph{Joint-embedding predictive architectures.} Joint-embedding predictive architectures (JEPAs) learn representations by predicting the embeddings of masked inputs rather than reconstructing pixels~\citep{ijepa}. V-JEPA extends this objective to video~\citep{vjepa}, V-JEPA~2 scales it~\citep{vjepa2}, and V-JEPA~2.1 improves the dense features of V-JEPA~2~\citep{vjepa21}, which makes it especially well suited to robotic policies. JEPA representations have since been used in robot policies: ACT-JEPA learns a JEPA-style latent space jointly with the policy~\citep{actjepa}, VLA-JEPA and JEPA-VLA bring V-JEPA targets or features into vision-language-action models~\citep{vlajepa,jepavla}, and JEPA-WAM and LeapBot-WA build world-action models on V-JEPA features~\citep{jepawam,leapbot}. These two world-action models are the closest to La-JEPA, but both initialize their predictors from pretrained vision-language or video generation models.

\paragraph{Compact policies.} Policies that are competitive without inheriting either a VLM or video backbone are scarce. While Diffusion Policy~\citep{diffusion_policy}, DP3~\citep{dp3} and ACT~\citep{act} exist, their performance is rarely close to VLAs and World Models. MINT predicts coarse-to-fine action tokens with a separately trained action tokenizer~\citep{mint} and reaches much better success rates, but remains below the strongest WAMs and VLAs. La-JEPA fits in this category as it trains its policy DiTs from scratch while adding WAM-style future prediction in a pretrained JEPA feature space.

\section{La-JEPA}
\label{sec:method}

\begin{figure}[t]
\centering
\includegraphics[width=\linewidth]{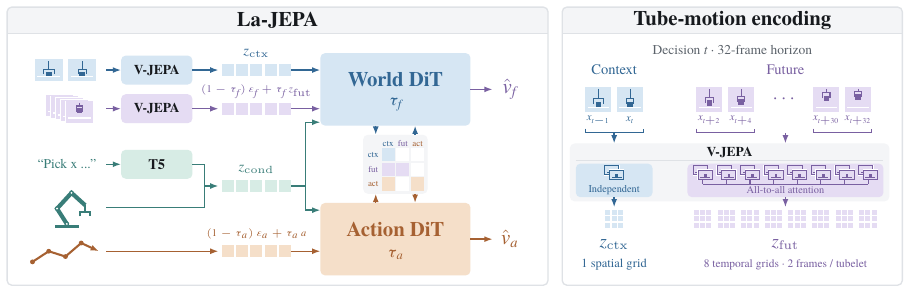}
\caption{La-JEPA. Left: general architecture. Right: context and future encoding with V-JEPA~2.1.}
\label{fig:architecture}
\end{figure}

La-JEPA learns action generation jointly with future-feature prediction in a pretrained V-JEPA~2.1 ViT-L representation space, following the separation of training-time world modeling from inference-time action generation in Fast-WAM~\citep{fastwam} and Faster-WAM~\citep{fasterwam}. At decision time $t$, the policy maps multi-camera observations $o_t$, proprioception $s_t$, and a language instruction $\ell$ to an action chunk $A=(a_t,\ldots,a_{t+H-1})$. This section describes the fixed architecture and training procedure (Figure~\ref{fig:architecture}); Section~\ref{sec:design-analysis} examines the choices within it.

\subsection{Architecture}
\label{sec:architecture}

\paragraph{Visual representations.} The pretrained V-JEPA~2.1 ViT-L encoder~\citep{vjepa21} processes each camera independently, producing observation features $z_{\mathrm{ctx}}$ from the most recent frames, and future targets $z_{\mathrm{fut}}$ representing the 32 frames ahead. We normalize and spatially reduce these feature grids before projecting them into the World DiT's embedding space. They are the model's only visual input and prediction target, so the model gets no separate RGB pathway. Future predictions are matched to targets directly in V-JEPA space and require no pixel reconstruction.

\paragraph{Language and proprioception.} A frozen FLAN-T5-large encoder~\citep{flan_t5} embeds the instruction $\ell$, and learned projections map proprioception $s_t$ to additional tokens; together they form the conditioning sequence $z_{\mathrm{cond}}$, which both DiTs read through cross-attention.

\paragraph{World--action architecture.} The model is made of two diffusion transformers (DiTs)~\citep{dit}: a World DiT over observation and future-feature tokens and an Action DiT over the action chunk. Both are initialized from scratch with separate attention projections and feed-forward layers, so pretrained knowledge enters only through the visual and language encoders. Rotary visual positional encodings~\citep{rope} preserve spatial and temporal information. The two DiTs interact at every layer, where action queries read the World DiT's observation keys and values alongside its own. The attention mask keeps future information out of the action pathway: observation tokens attend only to observations, future tokens to observations and futures, and action tokens to observations and actions. Actions therefore never access future features directly, yet predictive supervision still shapes the observation representations they read through the shared World DiT parameters during training.

\subsection{Training and Inference}
\label{sec:training-inference}

\paragraph{Joint flow matching.} We train both DiTs with future and action-space flow matching~\citep{flow_matching}. For each training example, we independently sample timesteps $\tau_f,\tau_a\sim\mathcal{U}(0,1)$ and Gaussian noise $\epsilon_f,\epsilon_a$, forming the noisy inputs
\begin{equation}
z_{\mathrm{fut}}^{\tau_f}=(1-\tau_f)\epsilon_f+\tau_f z_{\mathrm{fut}},
\qquad
A^{\tau_a}=(1-\tau_a)\epsilon_a+\tau_a A.
\label{eq:flow-interpolation}
\end{equation}
The model receives $(z_{\mathrm{ctx}},z_{\mathrm{fut}}^{\tau_f},A^{\tau_a},z_{\mathrm{cond}})$ and predicts velocities $\hat{v}_f$ and $\hat{v}_a$ under the attention mask as described above. The World DiT is conditioned on $\tau_f$ and the Action DiT on $\tau_a$. The training objective is
\begin{equation}
\mathcal{L}
=
\mathbb{E}\left[
\operatorname{MSE}(\hat{v}_f,z_{\mathrm{fut}}-\epsilon_f)
+
\operatorname{MSE}(\hat{v}_a,A-\epsilon_a)
\right],
\label{eq:training-objective}
\end{equation}
where each mean squared error averages over its supervised elements.

\paragraph{Action inference.} At inference, we drop the future-feature tokens, initialize the action chunk from Gaussian noise, and integrate the action velocity field from $\tau_a=0$ to $\tau_a=1$ with Euler steps, holding the World DiT's timestep conditioning at a fixed value $\tau_{\mathrm{ctx}}=0$ (since we do not predict the futures anyway). Since observation tokens attend only to one another under fixed conditioning, their per-layer keys and values are computed once and cached, and only the Action DiT runs at each denoising step. Like most recent policies, we convert the predicted action tokens to the robot's action space, execute the first $k$ actions of the chunk ($k=10$ on LIBERO and 24 on RoboTwin, as in Fast-WAM), and then predict a new chunk from the latest observations.

\section{Design Analysis}
\label{sec:design-analysis}

\begin{figure}[t]
\centering
\begin{minipage}[c]{0.56\linewidth}
\centering
\includegraphics[width=\linewidth]{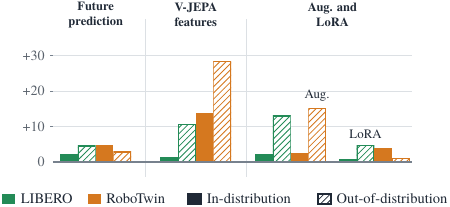}
\end{minipage}\hfill
\begin{minipage}[c]{0.41\linewidth}
\caption{Loss in success (points) when each of the three main components is removed from the final S recipe, all else fixed: future prediction (action-only training), V-JEPA features (replaced by Wan2.2 VAE latents), and image augmentation or encoder LoRA. Green: LIBERO; orange: RoboTwin~2.0; solid: in-distribution; hatched: out-of-distribution. All components, values, and 95\% intervals in Table~\ref{tab:design}.}
\label{fig:ablation-deltas}
\end{minipage}
\end{figure}

\paragraph{Experimental setup.}
We test the design choices behind Section~\ref{sec:evaluation} with the small (S) policy. Starting from the final recipe (Section~\ref{sec:final-recipe}), we remove one component at a time with data, schedule, seed, and evaluation fixed, so all blocks share a single reference: 85.6\% clean and 33.1\% randomized success on RoboTwin, 97.9\% on LIBERO, and 79.5\% on LIBERO-Plus. In-distribution evaluation uses LIBERO or clean RoboTwin scenes~\citep{libero,robotwin2}; out-of-distribution evaluation uses LIBERO-Plus~\citep{liberoplus} or randomized scenes. Figure~\ref{fig:ablation-deltas} reports the loss from removing each of the three main components in percentage points (hereafter, points); Table~\ref{tab:design} gives all components with absolute values and intervals, and Appendix~\ref{app:design-details} documents the protocols. Across three training seeds, the reference varies by 0.5 points on clean and 2.2 on randomized RoboTwin, 0.1 on LIBERO, and 0.4 on LIBERO-Plus.

\subsection{Prediction Target, Feature Space, and Adaptation}
\label{sec:predictive-design}
\label{sec:robustness-adaptation}

\paragraph{Predictive supervision.}
We first test whether predicting future features benefits the policy, comparing joint action and future-feature prediction with action-only training (Figure~\ref{fig:ablation-deltas}). Removing future prediction costs 4.8 points in clean and 2.9 in randomized RoboTwin scenes, 2.2 on LIBERO, and 4.5 on LIBERO-Plus, even though futures are never predicted at inference.

\paragraph{Feature space.}
We next ask in which space prediction should be learned, replacing frozen V-JEPA~2.1 features with latents of the VAE of Wan2.2~\citep{wan}, the video model underlying Fast-WAM, Faster-WAM, and LeapBot-WA (Figure~\ref{fig:ablation-deltas}). Normalization and projections are representation-specific, so the comparison covers each representation with its interface to the policy. The representation is the largest single factor: VAE latents cost 13.7 points in clean and 28.4 in randomized RoboTwin scenes, where the VAE policy succeeds in only 4.7\% of episodes, close to the video-generator WAMs in Table~\ref{tab:robotwin-comparison}, and 10.7 points on LIBERO-Plus against 1.4 on LIBERO. The robustness components (augmentation, encoder LoRA, and camera and proprioception dropout) do not compensate for the representation: with them, the VAE policy only matches a V-JEPA policy without them on LIBERO-Plus (68.9\% vs.\ 67.3\%). These results echo those of LDA-1B, which similarly reports a large gain from replacing VAE latents with DINO features~\citep{lda1b}.

\begin{table}[t]
\centering
\small
\renewcommand{\arraystretch}{1.1}
\caption{LIBERO and LIBERO-Plus success (\%) of policies without embodied pretraining (all use pretrained perception). Parameters in billions. LIBERO-Plus: episode-weighted success over all 10{,}030 variants with corrected or released instructions (Section~\ref{sec:libero-plus-text}); corrected-instruction scores exist only for the policies we re-evaluated. Best in bold, second best underlined unless the best is tied. La-JEPA values are means over three training seeds (Appendix~\ref{app:libero-revision}). $^{\S}$Scored at the upstream step limits used for all policies; with its own longer limits (400 steps, 700 for LIBERO-Long), NowWAM reports 87.7\% on the released instructions.}
\label{tab:main-comparison}
\setlength{\tabcolsep}{4pt}
\input{tables/manuscript_main_comparison_table.tex}
\end{table}

\begin{table}[t]
\centering
\small
\renewcommand{\arraystretch}{1.1}
\caption{RoboTwin~2.0 success of policies without embodied pretraining, trained on clean demonstrations only or on the full clean and randomized set, in clean and randomized scenes. Parameters in billions, for the RoboTwin configurations. Mean: equally weighted mean of clean and randomized success. Best in bold, second best underlined, within each training regime. Full-training La-JEPA values use one training seed with 5{,}000 episodes per condition; Faster-WAM parameters are those of its first version. Clean-training La-JEPA values are means over three training seeds, each evaluated on 5{,}000 episodes per condition (Appendix~\ref{app:libero-revision}).}
\label{tab:robotwin-comparison}
\setlength{\tabcolsep}{4pt}
\input{tables/manuscript_robotwin_comparison_table.tex}
\end{table}

\paragraph{Temporal encoding.}
V-JEPA~2.1 can also encode videos. \emph{Static} encodings treat every frame as an image; \emph{tube} encoding processes two-frame clips aligned with the encoder's tubelets, giving the observation a motion cue; \emph{tube-motion}, used in the final recipe, also jointly encodes the 16 future frames, so each future feature grid attends over the whole predicted horizon (Figure~\ref{fig:architecture}, right). Replacing tube-motion with tube or static encoding changes RoboTwin success by at most 1.1 points, within seed variation; on LIBERO, tube encoding matches tube-motion in distribution but loses 1.1 points on LIBERO-Plus (Table~\ref{tab:design}).

\paragraph{Token reduction.}
Before entering the World DiT, the feature grids are reduced spatially, by averaging neighboring features (mean pooling) or concatenating them into one token (patch grouping). Patch grouping improved randomized RoboTwin success, so the RoboTwin recipe uses it; LIBERO keeps mean pooling (89.35\% against 94.05\%; Appendix~\ref{app:token-reduction}).

\paragraph{Data augmentation and encoder adaptation.}
Finally, we combine image augmentation of the observations with LoRA~\citep{lora} adaptation of the observation encoder; future targets are still computed from unmodified frames by the original encoder for training efficiency purposes (Appendix~\ref{app:features}). Image augmentation brings most of the out-of-distribution robustness (Figure~\ref{fig:ablation-deltas}): removing it costs 13.0 points on LIBERO-Plus and 15.1 in randomized RoboTwin scenes, against 2.4 in-distribution. Removing encoder LoRA costs 4.7 points on LIBERO-Plus and 4.0 in clean RoboTwin scenes.

\subsection{Final Recipe}
\label{sec:final-recipe}
These studies define the final recipe: joint action and future prediction in V-JEPA~2.1 tube-motion features, and image augmentation with encoder LoRA. The recipe also drops cameras and proprioception during training, which mainly helps on LIBERO-Plus (4.0 and 1.0 points; Appendix~\ref{app:design-details}). We train it at three sizes, S, B, and L (Table~\ref{tab:protocol-settings}), which correspond to 0.10, 0.34, and 1.14\,B trainable parameters respectively, and at a fourth size, XL (2.14\,B), on the full RoboTwin set. Appendix~\ref{app:protocols} documents the remaining benchmark-specific settings.

\section{Experimental results}
\label{sec:evaluation}

\paragraph{Setup.}
We evaluate the final recipe against WAM and VLA policies that, like La-JEPA, use no embodied pretraining on robot data beyond the target demonstrations (Figure~\ref{fig:main-efficiency}, Tables~\ref{tab:main-comparison} and~\ref{tab:robotwin-comparison}), and against the closest latent world-action models (Table~\ref{tab:jepa-comparison}); we then compare computational cost and, on four physical tasks, real-robot success (Table~\ref{tab:real-robot}). Model sizes distinguish the parameters updated on target demonstrations (\emph{trainable}) from the full inventory including frozen encoders (\emph{deployed}; Appendix~\ref{app:param-accounting}).

\begin{table}[t]
\centering
\small
\renewcommand{\arraystretch}{1.1}
\caption{Latent world-action models. Embodied PT: pretraining on robot data beyond the target demonstrations. $^{\ddagger}$Reported by \citet{robustwam}. La-JEPA LIBERO and LIBERO-Plus values are means over three training seeds (Appendix~\ref{app:libero-revision}). Bottom: clean (C2C) and randomized (C2R) RoboTwin~2.0 success, both policies trained on clean demonstrations of JEPA-WAM's 20 tasks. Best in bold, second underlined.}
\label{tab:jepa-comparison}
\setlength{\tabcolsep}{3.2pt}
\input{tables/manuscript_jepa_comparison_table.tex}
\end{table}

\subsection{Benchmark Performance and Generalization}
\label{sec:benchmark-performance}

\begin{figure}[t]
\centering
\begin{minipage}[c]{0.5\linewidth}
\centering
\includegraphics[width=\linewidth]{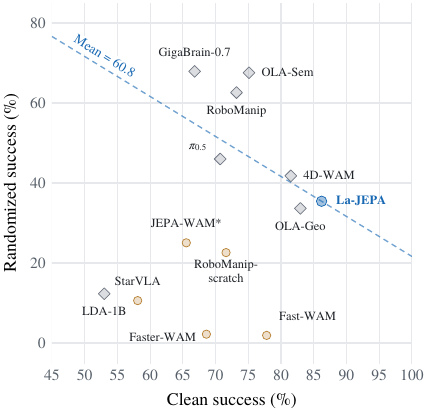}
\end{minipage}\hfill
\begin{minipage}[c]{0.46\linewidth}
\caption{RoboTwin clean versus randomized success of clean-trained policies. Blue: La-JEPA B; brown: no embodied pretraining; gray: embodied pretraining; RoboManip: Qwen-RobotManip. Points above the dashed line through La-JEPA have a higher mean of the two regimes. $*$: JEPA-WAM estimate (Appendix~\ref{app:protocols}); LDA-1B as reported by \citet{robustwam}.}
\label{fig:robotwin-clean-randomized}
\end{minipage}
\end{figure}
Both benchmarks test generalization: RoboTwin's randomized scenes add clutter, distractors, and new textures, and LIBERO-Plus perturbs the four LIBERO suites along seven axes such as camera viewpoint, lighting, and sensor noise. On RoboTwin, averaged over clean and randomized scenes, La-JEPA outperforms every baseline in both training regimes (Table~\ref{tab:robotwin-comparison}). On LIBERO-Plus, success increases with size. With the released instructions, L exceeds every baseline except NowWAM~\citep{nowwam}, which deploys four times more parameters; with corrected instructions (Section~\ref{sec:libero-plus-text}), all three sizes exceed or match every re-evaluated baseline, and L leads NowWAM by 1.8 points. On LIBERO, where most methods are close to saturation, all three sizes come within 0.6 points of the best baseline.

With clean-only training, RoboTwin success does not increase with size, as S, B, and L are within seed spread, and halving the action expert's width at L does not help either (86.6\% vs.\ 86.4\% clean, 32.8\% vs.\ 35.2\% randomized). The limit thus appears to be the training data, about 50 demonstrations per task: with the full RoboTwin training set, about ten times more data, \textbf{La-JEPA scales}, and success increases monotonically from S to a 2.1B-parameter XL, which reaches the highest mean among policies without embodied pretraining (93.7, against 93.4 for ImageWAM) with about three times fewer deployed parameters (Table~\ref{tab:robotwin-comparison}, Figure~\ref{fig:main-efficiency}).

Figure~\ref{fig:robotwin-clean-randomized} adds embodied-pretrained policies for reference. La-JEPA B's mean exceeds that of all but four, OLA-Sem, GigaBrain-0.7~\citep{gigabrain07}, Qwen-RobotManip, and 4D-WAM~\citep{fourdwam}, whose robot pretraining data covers part of the randomized variation.

\subsection{Instruction Leakage in LIBERO-Plus}
\label{sec:libero-plus-text}

\begin{table}[t]
\centering\footnotesize
\caption{Instruction received for two LIBERO-Plus variants. Bold: deviations from the training instruction.}
\label{tab:libero-plus-examples}
\setlength{\tabcolsep}{4pt}
\begin{tabular}{@{}>{\raggedright\arraybackslash}p{0.2\linewidth}>{\raggedright\arraybackslash}p{0.33\linewidth}>{\raggedright\arraybackslash}p{0.42\linewidth}@{}}
\toprule
Source & Camera viewpoint (LIBERO-Object) & Lighting (LIBERO-Spatial) \\
\midrule
Training, corrected (ours) & pick up the alphabet soup and place it in the basket & pick up the black bowl between the plate and the ramekin and place it on the plate \\
Released & \ldots{} in the basket \textbf{view 0 0 100 2 4 initstate 0} & \ldots{} place it on the plate \textbf{light 1} \\
Flex-$\pi$ & \textbf{Pick} the alphabet soup \ldots & \textbf{Pick the akita} black bowl \ldots \\
\bottomrule
\end{tabular}
\end{table}

The released LIBERO-Plus evaluator corrupts the instructions of 8{,}493 of its 10{,}030 variants by appending the perturbation's parameters (Table~\ref{tab:libero-plus-examples}); Flex-$\pi$~\citep{flexpi} noted this issue but substitutes BDDL annotations that paraphrase the training instructions for most variants, so that the language remains perturbed. We therefore also evaluate all open policies with corrected in-distribution instructions (Appendix~\ref{app:libero-plus-text}). The corruption affects video-generator-based WAMs most strongly, obscuring their true robustness to several perturbations: with corrected instructions, Fast-WAM gains 19.3 points, LeapBot-WA 8.1, and Cosmos Policy 4.1, whereas image-generator-based WAMs and VLM-based policies change by at most 1.4 points and La-JEPA gains 3.9 to 6.3. With corrected instructions, every La-JEPA size exceeds every re-evaluated policy without embodied pretraining (Table~\ref{tab:main-comparison}), and L leads NowWAM by 1.8 points with four times fewer deployed parameters; it also exceeds the embodied-pretrained $\pi_{0.5}$ and $\pi$-JEPA-WAM (Table~\ref{tab:libero-plus-text}).

\subsection{Comparison with Latent World-Action Models}
\label{sec:jepa-comparison}

La-JEPA also outperforms the two closest JEPA-based world-action models on LIBERO-Plus and RoboTwin~2.0, although both retain pretrained predictors (Table~\ref{tab:jepa-comparison}). LeapBot-WA, which also pretrains on robot data and deploys 10.4B parameters, falls below all three La-JEPA sizes on LIBERO-Plus with both the released and the corrected instructions (Table~\ref{tab:libero-plus-text} in Appendix~\ref{app:libero-plus-text}). JEPA-WAM, built on a pretrained Qwen2.5 model~\citep{qwen25}, is exceeded on LIBERO and LIBERO-Plus by all three La-JEPA sizes, S and B deploying fewer parameters. On RoboTwin, JEPA-WAM trains and evaluates on 20 of the 50 tasks. \textbf{Under this exact setting}, La-JEPA B trained on the same 20 tasks reaches 95.2\% clean and 56.5\% randomized success, ahead of JEPA-WAM by 15.3 and 19.6 points (Table~\ref{tab:jepa-comparison}, bottom); the additional tasks do not explain the gap, as training on all 50 tasks lowers B's randomized success on these 20 to 49.9\%. LDA-1B, pretrained on over 30k hours of human and robot data with a 4B vision-language model~\citep{lda1b}, remains far below on both benchmarks~\citep{robustwam}.

\subsection{Computational Cost}
\label{sec:capacity-efficiency}

From S to L, La-JEPA updates 0.10 to 1.14\,B parameters during training, far fewer than the video-generator and vision-language-action baselines (2.0 to 7.8\,B; Tables~\ref{tab:main-comparison} and~\ref{tab:robotwin-comparison}), while showing lower replan latency at matched or better performance; XL updates 2.14\,B, under half of the full-training RoboTwin baselines (4.5 to 6.0\,B). We time complete policy calls, with ten denoising steps on one consumer GPU (Table~\ref{tab:latency}, Appendix~\ref{app:latency}). Every policy is timed with the same degree of optimization, compiled builds on the same GPU ($\pi_{0.5}$: official JAX JIT); with these builds, S and B run two to five times faster than Fast-WAM and Faster-WAM, and S with original-kernel CUDA graphs, which reproduce uncompiled actions bit-for-bit, remains 2.6$\times$ faster than optimized Fast-WAM. Compiled builds do not significantly change closed-loop LIBERO success (Appendix~\ref{app:latency}), and the physical-robot policy runs compiled. Reducing inference to three denoising steps lowers latency further while keeping success within about one point (Appendix~\ref{app:eval-settings}), consistent with evidence that diffusion policies do not require many denoising steps~\citep{clemente2025twosteps}.

\begin{figure}[t]
\centering
\includegraphics[width=0.24\textwidth]{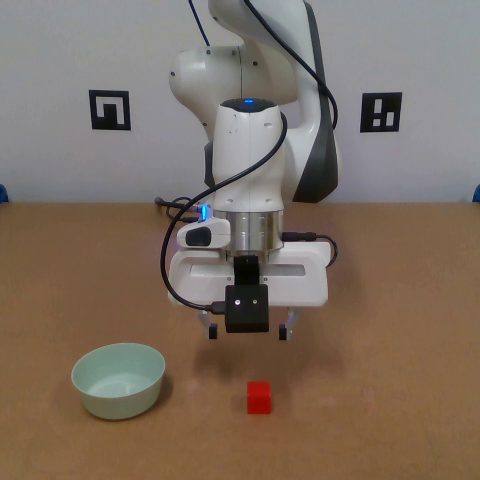}\hfill
\includegraphics[width=0.24\textwidth]{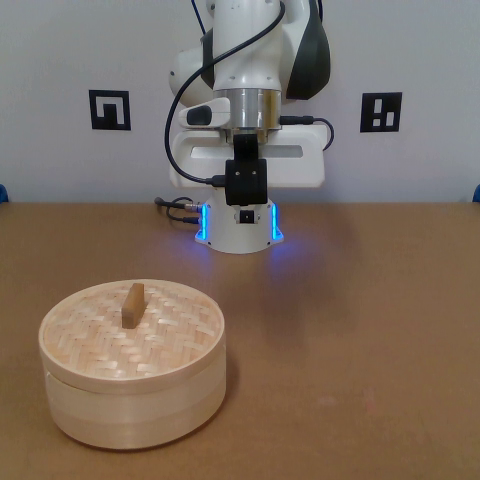}\hfill
\includegraphics[width=0.24\textwidth]{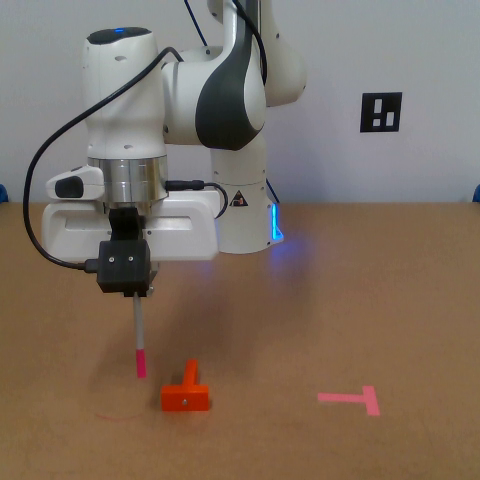}\hfill
\includegraphics[width=0.24\textwidth]{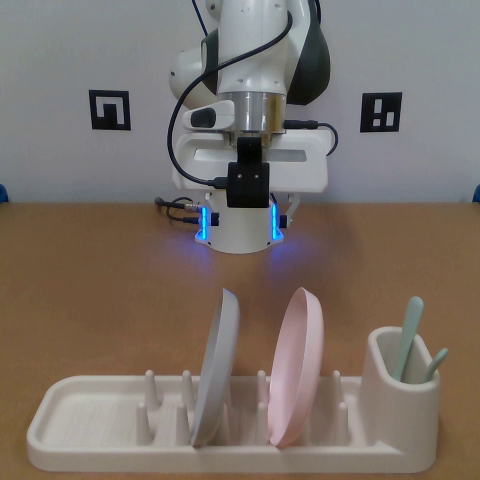}
\caption{Real-robot tasks: cube placement, fruit picking, Push T, and dish manipulation.}\label{fig:real-robot-tasks}
\end{figure}
\subsection{Real-Robot Experiments}
\label{sec:real-robot}

\begin{table}[t]
\centering
\caption{Real-robot successes out of 20 and benchmark policy-call latency in ms, not control-loop timing (Appendix~\ref{app:latency}); LingBot-VA's is rounded to 10\,ms. Position: full dish success with correct utensil placement; Average: mean over the four tasks. LingBot-VA dish: 10 trials. Bold: best.}
\label{tab:real-robot}
\footnotesize
\setlength{\tabcolsep}{4pt}
\input{tables/manuscript_real_robot_table.tex}
\end{table}

We train a La-JEPA B policy from scratch, with encoder LoRA but without image augmentation, on demonstrations of four Franka tabletop tasks (Figure~\ref{fig:real-robot-tasks}; Appendix~\ref{app:real_robot_setup}) and compare it with Fast-WAM, Faster-WAM, $\pi_{0.5}$, and LingBot-VA~\citep{lingbotva} in Table~\ref{tab:real-robot}. Cube placement is a control; fruit picking requires lifting a lid, setting it down, and choosing the correct fruit; Push T tests contact-heavy interactions; and the dish task requires placing two plates from a rack onto the table, then a spoon on the gray plate and a fork on the pink plate in one uninterrupted rollout, with interchangeable utensil positions that require distinguishing the utensils.

La-JEPA achieves the highest average success among policies without embodied pretraining and is comparable to the embodied-pretrained $\pi_{0.5}$, although its policy transformer is trained only on the target demonstrations. It is the most successful policy on Push T, whereas $\pi_{0.5}$ leads on the long-horizon dish task. On successful trials, La-JEPA also completes every task faster than the other policies without embodied pretraining (Appendix~\ref{app:real_robot_setup}), and faster than $\pi_{0.5}$ on fruit picking and Push T, though slower on the dish task. LingBot-VA rarely succeeds on Push T and dish manipulation, possibly needing more real-robot finetuning than the shared budget allowed, and is several times slower than every other policy.

\section{Limitations and Future Work}
\label{sec:discussion}

\paragraph{Scope of recipe optimization.}
We aim to show the potential of compact world-action policies rather than to optimize them exhaustively: beyond the targeted design studies, we did not systematically search learning rates, weight decay, dropout rates, LoRA configurations, or the relative capacity of the World and Action DiTs. Many strong baselines add external supervision objectives and robot-specific architectures; whether our improvements stack with them remains untested.

\paragraph{Large-scale pretraining.}
Our policies learn from target demonstrations alone, so how La-JEPA scales with embodied pretraining remains open. We consider this the principal future direction of research that should come out of this paper. La-JEPA remains below the strongest embodied-pretrained methods on randomized RoboTwin scenes with clean-only training, and its real-robot average, although comparable to that of $\pi_{0.5}$, comes from four tasks on one robot setup and does not establish parity across broader manipulation demands.

\section{Conclusion}
\label{sec:conclusion}
We asked whether world-action policies must inherit the architecture of a generative backbone, and whether a compact transformer trained from scratch can compete with those that do. Trained in a pretrained JEPA feature space, La-JEPA reaches the strongest RoboTwin~2.0 clean and randomized results among policies without embodied pretraining, and the strongest LIBERO-Plus results once the benchmark's corrupted instructions are corrected, while running up to 5$\times$ faster than Fast-WAM, and scales monotonically with size when trained on the full RoboTwin set. On four physical tasks, La-JEPA's average success is comparable to that of the embodied-pretrained $\pi_{0.5}$. Our ablation identifies the V-JEPA latent space, future prediction, and image augmentation as the key ingredients, and shows that heavy generative backbones are not required for high-performing policies.

\section*{AI Use Statement}
In this work, we used generative AI tools to implement methods, including the optimization of the training and dataloading pipeline which was particularly heavy, evaluation, and data-preprocessing code; to reformat datasets into precomputed feature stores (always through verifiable scripts); and to help design the infrastructure that enabled the interpretation of our large amount of results. We did not use generative AI tools to generate synthetic data, develop theoretical models, formulate or prove mathematical claims, or translate text; these uses are not applicable to this work. Additionally, we used generative AI tools to edit parts of the paper, to write the code that generates figures and tables from recorded results, to search for and summarize related literature, and to format references. We have reviewed all AI-assisted work: code was tested and validated against recorded outputs, reported numbers were traced to their evaluation records, figures and tables are generated by code from the underlying data directly, never by the AI tool itself, and references were checked against their arXiv entries. We take responsibility for the final content of this work, including text, claims, or artifacts produced with the aid of generative AI.

\section*{Ethics Statement}
This work uses public simulation benchmarks and robot demonstrations collected by the authors through teleoperation; it involves no human-subject studies and releases no new dataset. We do not foresee harms specific to this work beyond those of robot learning in general.

\section*{Reproducibility Statement}
Code for feature precomputation, training, and LIBERO, LIBERO-Plus, and RoboTwin~2.0 evaluation, with the configurations of the final S, B, and L policies, is available at \url{https://github.com/mt-clemente/lajepa}. Section~\ref{sec:method} specifies the architecture and training objective. Appendix~\ref{app:design-details} details the design studies, Appendix~\ref{app:protocols} the training configurations, evaluation protocols, and baseline sources, Appendix~\ref{app:latency} the latency and parameter accounting, Appendix~\ref{app:real_robot_setup} the real-robot setup, and Appendix~\ref{app:supplementary} supplementary results, including per-task evaluations.

\bibliography{references}
\bibliographystyle{iclr2027_conference}

\clearpage
\appendix

\section{Design Studies}
\label{app:design-details}

This appendix documents the design studies of Section~\ref{sec:design-analysis}. It first lists the values behind Figure~\ref{fig:ablation-deltas} and the setup shared by all studies, then gives the protocol of each comparison in the order of the main text.

\paragraph{Leave-one-out ablation.} Table~\ref{tab:design} lists all components, including the three shown in Figure~\ref{fig:ablation-deltas}. All rows use the S recipe, seed 42, and 150k updates; each removes one component from the final recipe with everything else fixed. RoboTwin policies train on the clean-only set (50 tasks) and are evaluated on 5{,}000 episodes per condition; LIBERO uses 2{,}000 episodes and LIBERO-Plus all 10{,}030 variants with the released instructions. Brackets give 95\% intervals over evaluation episodes for one training seed, where recorded; ns: not significant. Tube and static deltas on RoboTwin are within seed variation; the static variant was not run on LIBERO.

\paragraph{Camera and proprioception dropout.} To reduce reliance on any single input, we drop eligible cameras during training, always retaining at least one view and removing the dropped cameras' future targets from the loss, as in classifier-free guidance training~\citep{cfg}, and drop the proprioceptive conditioning with probability 0.1. Removing camera dropout costs 4.0 points on LIBERO-Plus and leaves the in-distribution benchmarks unchanged; its 2.8-point gain in randomized RoboTwin scenes is only 1.3 seed standard deviations. Removing proprioception dropout costs 1.0 point on LIBERO-Plus, with no other significant change (Table~\ref{tab:design}).

\paragraph{Additive design studies.} The recipe itself was chosen during development by additive studies, each comparing one choice with a baseline under its own reference recipe (Table~\ref{tab:design-additive}, blocks A to F). The paragraphs below document their protocols; comparisons in Table~\ref{tab:design-additive} hold only within a block.

\begin{table}[t]
\centering
\small
\renewcommand{\arraystretch}{1.1}
\caption{Leave-one-out ablation of the final S recipe: success (\%) with each component removed, and in parentheses the change in points against the reference, with 95\% intervals where recorded. $^{\dagger}$1.3 seed standard deviations; not established.}
\label{tab:design}
\setlength{\tabcolsep}{3pt}
\resizebox{\linewidth}{!}{\input{tables/ablation_loo_table.tex}}
\end{table}

\begin{table}[t]
\centering
\small
\renewcommand{\arraystretch}{1.1}
\caption{Additive design studies used during development: gain in success (points) of each design choice over its baseline; comparisons hold within a block.}
\label{tab:design-additive}
\setlength{\tabcolsep}{4pt}
\resizebox{\linewidth}{!}{\input{tables/ablation_deltas_table.tex}}
\end{table}

\paragraph{Shared setup.} All design studies use the original data and evaluation pipelines (Appendix~\ref{app:libero-revision}), so each comparison is made within that pipeline. All design-study transformers are trained from scratch and evaluated with EMA weights, including the corresponding EMA encoder weights for policies with encoder LoRA. Studies use the small (S) policy unless stated otherwise, and each result comes from a single training run, except for the three-seed LIBERO-Plus comparison of predictive supervision. Several studies train on a subset of the training demonstrations, nominally 95\%, which on LIBERO retains 1{,}663 of the 1{,}743 episodes; we call it the \emph{95\% split} and state the training data of each comparison below. LIBERO evaluations cover 2{,}000 episodes with a 700-step budget in every suite unless stated otherwise, and LIBERO-Plus evaluations all 10{,}030 episodes with the upstream suite-specific step limits; both redraw action noise at every replan. RoboTwin evaluation budgets are listed per block in Table~\ref{tab:design}; the 1{,}000-episode budgets are ablation evaluations, smaller than the 5{,}000-episode headline protocol (Appendix~\ref{app:protocols}).

\subsection{Prediction Target and Feature Space}
\label{app:ablations}

These studies determine whether the policy should predict future features, in which representation, and how motion and spatial resolution enter the tokens (Section~\ref{sec:predictive-design}).

\paragraph{Predictive supervision on LIBERO.} Joint action and future-feature prediction improves LIBERO-Plus success over action-only training for each of three training seeds, raising mean success from 74.35\% to 76.47\%, a gain of 2.12 points (Table~\ref{tab:objective-seeds}). Table~\ref{tab:design-additive} (block A) reports this mean gain; its in-distribution value uses the seed-42 pair at 150k updates, evaluated on 2{,}000 LIBERO episodes. Both objectives train on all 1{,}743 LIBERO episodes of the original pipeline for 150{,}000 updates with image augmentation and encoder LoRA, whereas the headline policies use the revised pipeline (Appendix~\ref{app:libero-revision}), so these controls are not directly comparable to Table~\ref{tab:main-comparison}. The action-only control omits future-feature supervision and computation while preserving the action pathway. For each objective and training seed, we select the checkpoint with the highest LIBERO success from the 30k, 60k, 90k, 120k, and 150k grid and evaluate it on all 10{,}030 LIBERO-Plus episodes with EMA weights, action-noise scale 1 redrawn at every replan, ten denoising steps, and ten-action execution chunks. Selection picks the same step for both objectives within each seed. Paired tests use identical task and evaluation-seed episodes for the two objectives.

\begin{table}[t]
\centering
\small
\caption{Predictive supervision across training seeds on LIBERO-Plus (success in \%; per-seed exact McNemar tests).}
\label{tab:objective-seeds}
\input{tables/objective_seed_comparison_table.tex}
\end{table}

\paragraph{Predictive supervision on RoboTwin.} On 5{,}000 episodes per regime, each policy on its own set of evaluation episodes, joint training achieves 80.16\% clean and 39.02\% randomized success, against 73.14\% and 35.34\% for action-only training. Both policies use the final augmentation recipe (Table~\ref{tab:protocol-augmentation}), all 2{,}500 clean demonstrations, 150{,}000 training updates, and a global batch of 400, split into four data-parallel ranks of 100 for joint training and eight of 50 for action-only training; the action-only policy neither loads nor computes future targets. Selection on the 1{,}000-episode clean grid picks 150k for joint training and 120k for action-only training, with ties broken toward the later step.

\paragraph{Feature space.} Block B compares frozen, static V-JEPA~2.1 features with Wan2.2 VAE latents at 150k updates, with action and future-feature prediction on the 95\% split. Within each benchmark, feature dimensions, normalization, and input and output projections remain representation-specific, so the comparison concerns the two representation systems rather than their pretraining objectives in isolation. On RoboTwin, both policies use 384-dimensional, 12-layer transformers with patch grouping for observations and future targets and are evaluated on 5{,}000-episode clean and randomized manifests with 24-action execution chunks. On LIBERO, both use the 768-dimensional, 12-layer B transformer with $2\times$ spatial mean pooling and 10-action execution chunks; LIBERO evaluation instead uses a 600-step budget per suite and 30 settling steps, and LIBERO-Plus evaluation uses action-noise scale 1. On LIBERO-Plus, V-JEPA succeeds in 5{,}773 episodes and Wan2.2 in 5{,}173. All evaluations use EMA weights, a single observation context, and 10 denoising steps.

\paragraph{Additional feature-space comparisons.} Two further RoboTwin comparisons pool the future targets, each evaluated on 5{,}000 episodes per regime (clean only for the first). With full-data, static-feature training and patch-grouped observations, V-JEPA and Wan2.2 achieve 61.50\% and 61.44\% clean success at 150k. In the second, with 768-dimensional transformers, Wan2.2 leads on clean success (63.90\% versus 61.12\%), whereas V-JEPA leads on randomized success (15.00\% versus 1.28\%). V-JEPA's advantage therefore holds in randomized scenes but not consistently in clean ones.

\paragraph{Temporal encoding.} Block C compares three frozen-encoder policies with patch grouping, trained on the 95\% split, that differ only in the encoding: static, tube, or tube-motion. Each regime uses 1{,}000 episodes on the same scenes, and randomized success is read at the checkpoint selected on clean episodes: 150k for static and tube-motion, and 120k for tube.

\paragraph{Token reduction.}
\label{app:token-reduction}
On RoboTwin, grouping neighboring features into one token by concatenation (patch grouping) instead of averaging them (mean pooling) changes clean success by $+0.3$ points and raises randomized success by 4.1 points. Both policies apply the reduction to two-frame observations and to motion-based future targets and are compared at 150k updates; each clean and randomized evaluation uses 1{,}000 episodes, 32 samples per pixel, rendering depth 8, OIDN denoising, and action noise fixed within each episode. The pooling policy uses the full training set and the patch-grouping policy the 95\% split, and their target normalization and projections also differ, so the comparison does not isolate the reduction operator. By contrast, on LIBERO, in a one-variable comparison between stored-feature S policies with tube-motion encoding at 150k updates, evaluated identically on 2{,}000 episodes, mean pooling reaches 94.05\% against 89.35\% for patch grouping. The final recipe uses patch grouping on RoboTwin, reducing the head camera $2\times$ and the wrist cameras $4\times$, and $2\times$ mean pooling on LIBERO (Appendix~\ref{app:features}). Appendix~\ref{app:future-pooling} reports a related sweep that pools only the future targets.

\subsection{Robustness and Encoder Adaptation}
\label{app:robustness-details}

These studies target robustness to changes in the observations through camera and proprioception dropout, image augmentation, and encoder adaptation (Section~\ref{sec:robustness-adaptation}). All LIBERO variants in blocks D and F train on the 95\% split and are selected on LIBERO success: stored-feature policies with a frozen encoder at 90k updates, and online-encoder variants at 120k.

\paragraph{Camera dropout on LIBERO.} The LIBERO row of block D compares either-camera with wrist-only dropout in stored-feature policies. Allowing either camera to be dropped changes both camera eligibility and the expected number of missing streams, from 0.10 to 0.19 per example; at least one stream is always retained. Table~\ref{tab:libero-perturbations} reports the effect of this change, and of the adaptation variants below, on the camera-viewpoint and sensor-noise categories of LIBERO-Plus.

\begin{table}[t]
\centering
\caption{LIBERO-Plus success under camera-viewpoint and sensor-noise perturbations (\%).}
\label{tab:libero-perturbations}
\input{tables/manuscript_libero_perturbations_table.tex}
\end{table}

\paragraph{Camera and proprioception dropout on RoboTwin.} The RoboTwin row of block D compares two frozen-encoder, static-feature policies at 150k updates that differ only in wrist-camera dropout with probability 0.1. The proprioception-dropout row of block E compares two policies of the same family that differ only in proprioception dropout with probability 0.1, read at each policy's best checkpoint with an evaluation implementation different from that of the camera-dropout row. Both comparisons use 1{,}000 episodes per regime.

\paragraph{Augmentation and encoder adaptation.} On LIBERO, block F compares encoder LoRA, image augmentation, and both combined against the plain, either-camera-dropout policy of block D; the variants run the encoder online and are selected at 120k updates, whereas the plain reference uses stored features and is selected at 90k. On RoboTwin, the frozen-encoder policy without augmentation is compared with the augmentation-and-LoRA policy trained with the final augmentation pipeline, both at 150k updates and evaluated on 5{,}000 episodes per regime; the plain policy uses 2{,}400 of the 2{,}500 demonstrations and the augmentation-and-LoRA policy all 2{,}500, and the evaluation implementations also differ. Because the policies also differ in training duration on LIBERO and in training data on RoboTwin, block F measures recipe-level gains rather than isolated effects.

\section{Training and Evaluation Protocols}
\label{app:protocols}

This appendix specifies the headline training recipe, the data and checkpoint selection behind the reported La-JEPA results, the benchmark protocols, and the sources of baseline results.

\paragraph{Architecture and optimization.} Table~\ref{tab:protocol-settings} summarizes the architecture, optimization, conditioning, and inference settings of the headline recipe at the three policy sizes.

\begin{table}[t]
\centering\small
\caption{Architecture, training, and inference settings.}
\label{tab:protocol-settings}
\setlength{\tabcolsep}{3pt}
\renewcommand{\arraystretch}{1.12}
\input{tables/manuscript_protocol_settings_table.tex}
\end{table}

\paragraph{Observation augmentation.} Table~\ref{tab:protocol-augmentation} lists the final augmentation recipe. After the gate that leaves a window unmodified, transformation gates are independent, and parameters are shared across frames. Only observations are augmented; future targets are computed from unmodified frames by the frozen encoder.

\begin{table}[t]
\centering\small
\caption{Observation augmentation. Probabilities are conditional on not bypassing augmentation; fog is one of three corruption types.}
\label{tab:protocol-augmentation}
\setlength{\tabcolsep}{3pt}
\renewcommand{\arraystretch}{1.12}
\input{tables/manuscript_protocol_augmentation_table.tex}
\end{table}

\paragraph{Precomputed features and token reduction.}
\label{app:features}
Before training, we encode every demonstration once with the frozen V-JEPA~2.1 encoder and store the resulting features. Stored-feature policies, used in most design studies, read both observations and future targets from these stores and never run the encoder during training. Encoder LoRA and image augmentation require encoding the observations at every training step, so the final policies encode their observation clips online but still read their future targets from the stores. Encoding the targets online as well would dominate the encoder cost of each step, since each camera's targets span 16 frames, eight times as many encoder tokens as its two-frame observation clip. Targets from the adapting encoder also do not improve success (Appendix~\ref{app:adapted-targets}), so the targets remain those of the frozen encoder.

The features are then reduced spatially before entering the World DiT. Mean pooling, the simplest reduction, averages each $2\times2$ neighborhood of a feature grid into one token. Because it can be applied once when the features are precomputed, it shrinks the stored features fourfold. Most of our development therefore used pre-pooled stores, which are much smaller and far easier to load during training than full-resolution V-JEPA features. Patch grouping instead concatenates neighboring features into one wider token, reducing the number of tokens without discarding any feature values, but the stored features keep their full size. Mean pooling suffices on LIBERO, whereas on RoboTwin the detail it averages away matters: patch grouping raises randomized success by 4.1 points (Appendix~\ref{app:token-reduction}), so the final RoboTwin recipe uses patch grouping.

\paragraph{Training data and checkpoint selection.} On LIBERO, all three sizes train on all 1{,}777 demonstrations retained by the revised pipeline (Appendix~\ref{app:libero-revision}) for 150{,}000 updates, and we report the final checkpoints' performances averaged over three training seeds; L additionally uses full-depth LoRA. The final RoboTwin S, B, and L policies train on the 2{,}438 clean demonstrations retained by the revised pipeline (Appendix~\ref{app:libero-revision}) with three training seeds each. For each seed, the checkpoint is selected among 90k, 120k, and 150k updates on a 1{,}000-episode clean grid disjoint from the evaluation episodes, which picks 150k for every seed except one L seed (120k); the selected policies are evaluated on 5{,}000 episodes per condition with action noise redrawn at every replan, and we report the mean over seeds.

\paragraph{Benchmark protocols.} LIBERO scores are success rates over the four standard suites. LIBERO-Plus scores are success rates over all 10{,}030 evaluation episodes of the seven perturbation categories, including background textures, weighting every episode rather than every category equally. Literature aggregates are recomputed from reported per-category rates when needed. La-JEPA's headline LIBERO and LIBERO-Plus evaluations use MuJoCo~3.3.2 and the upstream suite-specific step limits (220, 280, 300, and 520 steps, after 10 settling steps). The design studies, evaluation-setting sweeps, and compiled-build validation keep a flat budget in every suite, 700 steps except for the LIBERO pair of block B (600 steps after 30 settling steps; Appendix~\ref{app:ablations}), so each comparison is made within one protocol. RoboTwin policies are trained on clean demonstrations and evaluated in clean and randomized scenes, and the mean of the two regimes is the leaderboard aggregate. La-JEPA's RoboTwin policies are conditioned on each episode's own RoboTwin instruction in both training and evaluation. JEPA-WAM reports 20 RoboTwin tasks, so Table~\ref{tab:jepa-comparison} restricts our evaluations to the same tasks, using 2{,}000 episodes per condition and training seed. JEPA-WAM trains only on these 20 tasks, whereas our policies train on all 50. The subset is easier than the full benchmark: per-task DP3 and ACT policies reach 73.9\% and 51.8\% clean success on it~\citep{jepawam}, against 55.2\% and 29.7\% over all 50 tasks~\citep{robotwin2}. Projected to all 50 tasks, JEPA-WAM's success is estimated at 65.5\% clean and 25.0\% randomized (see the full-suite estimate below), below every La-JEPA size in Table~\ref{tab:robotwin-comparison}. Methods pretrained on robot trajectories beyond the target demonstrations are outside the main comparison and are explicitly identified in the latent world-action model comparison, the leaderboard figure, and the physical-robot comparison.

\paragraph{Baseline sources.} Baseline scores come from the following sources.
\begin{itemize}[leftmargin=*,itemsep=1pt,topsep=2pt,parsep=0pt]
\item \textbf{Re-evaluations of released checkpoints:} All LIBERO-Plus scores of StarVLA~\citep{starvla}, JEPA-WAM, Fast-WAM, Cosmos Policy, ImageWAM, NowWAM, LeapBot-WA, $\pi_{0.5}$, and $\pi$-JEPA-WAM, with both the released and the corrected instructions (Section~\ref{sec:libero-plus-text} and Appendix~\ref{app:libero-plus-text}), and the LIBERO scores of StarVLA and NowWAM. For StarVLA, whose variant the Qwen-RobotManip report does not specify, we evaluate the released LIBERO checkpoint with a Qwen3-VL-4B backbone and a flow-matching action head; for LeapBot-WA, we evaluate its LIBERO-trained checkpoint, matching the zero-shot LIBERO-Plus transfer reported in its paper.
\item \textbf{Training:} Faster-WAM reports RoboTwin results only for training on clean and randomized demonstrations, so we train it on the clean demonstrations ourselves, adapting the open-source Fast-WAM code as described in the Faster-WAM paper.
\item \textbf{Qwen-RobotManip report~\citep{qwen_robotmanip}:} All Qwen-RobotManip-scratch scores and the RoboTwin scores of StarVLA, as well as the embodied-pretrained Qwen-RobotManip model in Figure~\ref{fig:robotwin-clean-randomized}, taken from its clean-to-randomized comparison.
\item \textbf{RoboTwin~2.0 leaderboard and report~\citep{robotwin2}:} Fast-WAM's clean-trained RoboTwin scores and the embodied-pretrained policies added in Figure~\ref{fig:robotwin-clean-randomized} come from the official leaderboard\footnote{\url{https://robotwin-platform.github.io/leaderboard}}: 4D-WAM~\citep{fourdwam}, GigaBrain-0.7~\citep{gigabrain07}, and OLA-Sem and OLA-Geo, which are leaderboard entries without an accompanying paper. The ACT and DP3 RoboTwin scores in Figure~\ref{fig:main-efficiency-trainable} come from the benchmark report.
\item \textbf{Method papers:} All other LIBERO and LIBERO-Plus scores, and JEPA-WAM's 20-task RoboTwin scores, come from each method's paper; in Figure~\ref{fig:main-efficiency-trainable}, SmolVLA's LIBERO scores are those of its variants without robot pretraining~\citep{smolvla}, and LDA-1B's scores in Table~\ref{tab:jepa-comparison} are those reported by \citet{robustwam}. Where a paper prints a category-averaged LIBERO-Plus score (Faster-WAM and MINT), we recompute the episode-weighted score from its per-category rates, so our values can differ from the printed averages.
\end{itemize}
LeapBot-WA's clean-plus-randomized training regime is excluded from the RoboTwin comparison of Table~\ref{tab:jepa-comparison}. In Figure~\ref{fig:robotwin-clean-randomized}, the La-JEPA point is B, and the dashed contour fixes the arithmetic mean of clean and randomized success.

\paragraph{Data and evaluation pipelines:}
\label{app:libero-revision}
The headline policies use the data and evaluation pipelines below, with three training seeds per size. The design studies, representation diagnostics, evaluation-setting sweeps, adapted-target comparison, and compiled-build validation use the original version of these pipelines, with the same architecture, objective, and augmentation strengths; Tables~\ref{tab:libero-revision} and~\ref{tab:robotwin-revision} compare the two.

\emph{LIBERO:}
\begin{enumerate}[leftmargin=*,itemsep=1pt,topsep=2pt,parsep=0pt]
\item \textbf{Simulator version:} Demonstrations are rendered and all policies are evaluated with MuJoCo~3.3.2, the version used by the compared baselines; under newer releases such as 3.9.0, the bowl in the task ``pick up the black bowl on the ramekin'' is unseated at reset.
\item \textbf{Demonstration filtering:} Each demonstration's recorded actions are replayed from its recorded initial state, and only demonstrations that end in task success are kept: 1{,}777 of the 2{,}000.
\item \textbf{Proprioception alignment:} Each frame is paired with the proprioceptive state at its own time step, as at evaluation.
\item \textbf{Normalization:} States and actions use per-channel 1st/99th-percentile scaling to $[-1,1]$, as in OpenVLA-OFT~\citep{openvla_oft} and $\pi_{0.5}$~\citep{pi05}.
\item \textbf{End-of-episode windows:} Training windows extend to the end of each demonstration, and missing future targets are masked from the loss.
\item \textbf{Training schedule:} All sizes train on all retained demonstrations for 150{,}000 updates, and we report the final checkpoint averaged over the three seeds.
\item \textbf{Evaluation protocol:} Each episode takes 10 settling steps before the policy acts and is scored at the upstream per-suite step limits (220, 280, 300, and 520 steps).
\end{enumerate}
Table~\ref{tab:libero-revision} compares these policies with the original-pipeline ones, which were evaluated under MuJoCo~3.9.0.

\begin{table}[t]
\centering\small
\caption{Revised versus original LIBERO policies (success in \%). Revised: final 150k checkpoint, mean over three training seeds $\pm$ sample standard deviation across seeds; LIBERO-Plus categories are three-seed means. Original: one training run at 120k updates. Both use the upstream suite-specific step limits; the original LIBERO values are counted from recorded rollouts without settling steps, whereas the revised evaluations take 10 settling steps before the policy acts.}
\label{tab:libero-revision}
\setlength{\tabcolsep}{3pt}
\resizebox{\linewidth}{!}{%
\begin{tabular}{lrrrrrrrrrrr}
\toprule
 & \multicolumn{7}{c}{LIBERO-Plus categories (revised)} & \multicolumn{2}{c}{LIBERO-Plus} & \multicolumn{2}{c}{LIBERO} \\
\cmidrule(lr){2-8}\cmidrule(lr){9-10}\cmidrule(lr){11-12}
Size & Camera & Robot & Language & Light & Backgr. & Noise & Layout & Revised & Orig. & Revised & Orig. \\
\midrule
S & 80.4 & 58.3 & 89.1 & 95.5 & 66.0 & 89.6 & 80.3 & 80.0 $\pm$ 0.4 & 75.2 & 97.9 $\pm$ 0.1 & 93.4 \\
B & 85.0 & 58.0 & 91.0 & 96.7 & 71.4 & 93.2 & 82.1 & 82.5 $\pm$ 0.4 & 78.9 & 98.0 $\pm$ 0.2 & 96.4 \\
L & 86.4 & 61.4 & 92.1 & 97.4 & 74.9 & 94.6 & 82.1 & 84.1 $\pm$ 0.4 & 82.3 & 98.1 $\pm$ 0.2 & 97.0 \\
\bottomrule
\end{tabular}}
\end{table}

\emph{RoboTwin~2.0:}
\begin{enumerate}[leftmargin=*,itemsep=1pt,topsep=2pt,parsep=0pt]
\item \textbf{Normalization:} States and actions use the same per-channel percentile scaling as on LIBERO, clamped to $\pm5$, with gripper commands mapped from $[0,1]$ to $[-1,1]$.
\item \textbf{Joint-angle wrapping:} In 62 of the 2{,}500 clean demonstrations, spread over 21 of the 50 tasks, the planned joint path crosses $\pm\pi$, where the wrap-around makes the joint trajectory discontinuous. We exclude these demonstrations and train on the remaining 2{,}438 (538{,}662 frames, at least 41 per task).
\item \textbf{End-of-episode windows:} Training windows extend to the end of each episode; padded steps and padded future-target rows are masked from the loss, as on LIBERO.
\item \textbf{Instructions:} Each episode is conditioned on its own official RoboTwin instruction.
\item \textbf{Task isolation:} Each task is evaluated in its own simulator process. In \texttt{dump\_bin\_bigbin} and \texttt{put\_bottles\_dustbin}, the upstream scene setup shifts the clutter region at every reset, so that it drifts across episodes; we keep this upstream behavior for comparability with leaderboard results, and the isolation prevents it from affecting other tasks, all of which pass a per-reset scene check.
\end{enumerate}
Table~\ref{tab:robotwin-revision} compares these policies with the original-pipeline ones: clean success rises by 5 to 8 points at every size, randomized success falls by 5 to 9 points, and the mean of the two regimes changes by at most one point.

\begin{table}[t]
\centering\small
\caption{Revised versus original RoboTwin~2.0 policies (clean-trained; success in \%, 5{,}000 episodes per condition). Revised: mean over three training seeds $\pm$ sample standard deviation across seeds. Original: one training run.}
\label{tab:robotwin-revision}
\begin{tabular}{lrrrrrr}
\toprule
 & \multicolumn{2}{c}{Clean} & \multicolumn{2}{c}{Randomized} & \multicolumn{2}{c}{Mean} \\
\cmidrule(lr){2-3}\cmidrule(lr){4-5}\cmidrule(lr){6-7}
Size & Revised & Orig. & Revised & Orig. & Revised & Orig. \\
\midrule
S & 85.5 $\pm$ 0.5 & 80.2 & 33.9 $\pm$ 2.2 & 39.0 & 59.7 $\pm$ 1.3 & 59.6 \\
B & 86.2 $\pm$ 0.5 & 78.8 & 35.4 $\pm$ 0.6 & 44.7 & 60.8 $\pm$ 0.5 & 61.8 \\
L & 86.0 $\pm$ 0.4 & 78.4 & 32.1 $\pm$ 2.8 & 39.1 & 59.0 $\pm$ 1.6 & 58.7 \\
\bottomrule
\end{tabular}
\end{table}

\paragraph{LIBERO-Plus instruction texts.}
\label{app:libero-plus-text}
Table~\ref{tab:libero-plus-texts} compares the instruction a policy receives under four sources. The LIBERO training demonstrations store each task's instruction in their metadata, identical to the instruction that standard LIBERO evaluation derives from the task's file name. The released LIBERO-Plus evaluator derives the instruction from the variant's file name instead, so every non-language variant receives the training instruction followed by the perturbation's parameters. Flex-$\pi$~\citep{flexpi} replaces it with the \texttt{:language} annotation of the variant's BDDL file, falling back to the base task's file for camera, robot-initial-state, and sensor-noise variants. These annotations are written independently of the training instructions: across the 8{,}493 non-language variants, only 2{,}351 match the training instruction exactly and 3{,}148 up to capitalization, so most variants still receive a paraphrase that the policy never saw during training. Our corrected text gives every non-language variant its base task's training instruction verbatim. For the 1{,}537 language variants, whose paraphrases are the intended perturbation, the released and corrected texts are identical, and Flex-$\pi$'s annotations contain the same words but differ in capitalization for 1{,}209 of them. All re-evaluations share one protocol; NowWAM, Fast-WAM, and ImageWAM, whose released evaluations use longer step limits, are scored at the upstream per-suite limits like every other policy. Table~\ref{tab:libero-plus-text} reports both evaluations. Because the corruption affects policies to different degrees, released-instruction scores partly rank policies by their sensitivity to corrupted text rather than by their robustness to the intended perturbation: correcting the instructions reverses the order of NowWAM and La-JEPA~B and~L, and of Cosmos Policy and La-JEPA~S. With corrected instructions, S is on par with NowWAM, and L's 1.8-point lead exceeds both its seed-to-seed standard deviation (0.3 points) and the binomial standard error of a single 10{,}030-episode evaluation (about 0.35 points).

\begin{table}[t]
\centering\small
\caption{Instruction sources for LIBERO-Plus. Example: the first LIBERO-Spatial task under the first lighting perturbation. Match: non-language variants whose instruction equals the training instruction, out of 8{,}493.}
\label{tab:libero-plus-texts}
\setlength{\tabcolsep}{4pt}
\begin{tabular}{@{}>{\raggedright\arraybackslash}p{0.2\linewidth}>{\raggedright\arraybackslash}p{0.56\linewidth}r@{}}
\toprule
Source & Example instruction & Match \\
\midrule
LIBERO training data & pick up the black bowl between the plate and the ramekin and place it on the plate & --- \\
Released LIBERO-Plus & pick up the black bowl between the plate and the ramekin and place it on the plate \textbf{light 1} & 0 \\
Flex-$\pi$ (BDDL annotation) & \textbf{Pick the akita} black bowl between the plate and the ramekin and place it on the plate & 2{,}351 \\
Ours (corrected) & pick up the black bowl between the plate and the ramekin and place it on the plate & 8{,}493 \\
\bottomrule
\end{tabular}
\end{table}

\begin{table}[t]
\centering\small
\caption{LIBERO-Plus success (\%) with the released and the corrected instructions, all evaluated in our harness on all 10{,}030 variants (MuJoCo~3.3.2, seed 0, first initial state, 10 settling steps, upstream per-suite step limits). $\Delta$: corrected minus released. Best in bold, second best underlined. La-JEPA values are means over three training seeds. $^{\S}$One variant cannot be instantiated by LIBERO-Plus, so $n=10{,}029$. $^{\P}$Run-to-run variation of about $\pm0.5$ points. $\pi$-JEPA-WAM: $\pi_{0.5}$ trained with the auxiliary JEPA objective of JEPA-WAM~\citep{jepawam}.}
\label{tab:libero-plus-text}
\begin{tabular}{llrrr}
\toprule
Policy & Embodied PT & Released & Corrected & $\Delta$ \\
\midrule
La-JEPA S (ours) & no & 80.0 & 86.3 & $+6.3$ \\
La-JEPA B (ours) & no & 82.5 & \underline{87.3} & $+4.9$ \\
La-JEPA L (ours) & no & \underline{84.1} & \textbf{88.0} & $+3.9$ \\
NowWAM & no & \textbf{86.2} & 86.2 & $+0.0$ \\
Cosmos Policy$^{\S}$ & no & 80.8 & 84.9 & $+4.1$ \\
ImageWAM & no & 79.7 & 80.4 & $+0.7$ \\
JEPA-WAM$^{\S}$ & no & 77.1 & 77.8 & $+0.7$ \\
StarVLA$^{\S}$ & no & 76.7 & 77.6 & $+0.9$ \\
Fast-WAM & no & 51.5 & 70.8 & $+19.3$ \\
\midrule
$\pi_{0.5}$$^{\S\P}$ & yes & 84.3 & 85.7 & $+1.4$ \\
$\pi$-JEPA-WAM$^{\S\P}$ & yes & \underline{84.5} & 85.2 & $+0.6$ \\
LeapBot-WA$^{\S}$ & yes & 71.7 & 79.7 & $+8.1$ \\
\bottomrule
\end{tabular}
\end{table}

\paragraph{Action-noise scale for L.}
\label{app:action-noise}
All headline results use the default action-noise scale of 1, which multiplies the initial action noise at each replan; for the original-pipeline L policy (Appendix~\ref{app:libero-revision}), this gives 82.26\% on LIBERO-Plus. Appendix~\ref{app:eval-settings} reports a sweep over scales $\{0,0.25,0.5,1\}$ for that policy, in which scale 0.5 reaches 82.50\%; no scale gives a statistically supported improvement, so we report the default. The scale does not change the trained model or its parameter count.

\paragraph{JEPA-WAM full-suite estimate.} For the asterisk-marked points in Figures~\ref{fig:main-efficiency} and~\ref{fig:robotwin-clean-randomized}, we fit separate linear regressions from 20-task to 50-task RoboTwin success using seven policies with both scores: Diffusion Policy~\citep{diffusion_policy}, ACT~\citep{act}, RDT~\citep{rdt}, $\pi_0$~\citep{pi0}, and DP3~\citep{dp3} from the official leaderboard's clean-trained per-task results, and La-JEPA S and B. With success expressed in percent, the fits are $\hat{s}_{\mathrm{clean}}=-33.09+1.234s_{20}$ and $\hat{s}_{\mathrm{rand}}=-1.14+0.710s_{20}$, with leave-one-out mean absolute errors of 2.1 and 1.0 points, respectively. Applied to JEPA-WAM's reported 79.9\% clean and 36.9\% randomized success~\citep{jepawam}, they give estimates of 65.5\% and 25.0\% (mean 45.3\%). Figure~\ref{fig:main-efficiency} shows the mean estimate at 1.35\,B deployed parameters; these cross-policy estimates are excluded from the result tables.

\section{Inference Cost and Model Size}
\label{app:latency}

This appendix defines how we time policy calls and count parameters, the two cost measures of Section~\ref{sec:capacity-efficiency}.

\subsection{Latency}

\begin{table}[t]
\centering
\small
\caption{Policy-call latency (ms): batch size one, median of 100 calls after 20 warm-ups on one consumer GPU. OOM: out of memory; dashes: not measured. The timed L uses last-eight-block rather than full-depth LoRA, which does not change latency.}
\label{tab:latency}
\input{tables/manuscript_latency_table.tex}
\end{table}

\paragraph{Protocol.} Tables~\ref{tab:latency} and~\ref{tab:real-robot} time complete benchmark policy calls, not executed actions or control-loop cycles. Each timed call includes fresh instruction tokenization and encoding, input preparation, transfer, and host-side action output; loading, compilation, simulation, and transport are excluded. Text and vision encoders run on parallel streams where supported. Eager execution is uncompiled; original-kernel CUDA graphs reduce launch overhead without changing kernels; optimized builds also compile the computation. The head-to-head WAM rows use ten denoising steps and 32 predicted actions, whereas the three-step rows and $\pi_{0.5}$ (official JAX JIT, ten steps, native ten-action horizon) are separate operating points. Each method keeps its native inputs: La-JEPA uses two $256\times256$ cameras with two adjacent frames and FLAN-T5-large, timed in FP32 (our success evaluations use FP16), whereas in our setup Fast-WAM and Faster-WAM use two $224\times224$ single-frame cameras and BF16 UMT5-XXL~\citep{umt5}. Action noise is pre-drawn and copied inside the timed call for La-JEPA, and drawn on the CPU inside the call for Fast-WAM. LingBot-VA is timed on the same device under the same protocol with its native implementation and appears, rounded to 10\,ms, only in Table~\ref{tab:real-robot}.

\paragraph{Execution variants.} For La-JEPA, original-kernel CUDA graphs reproduce eager actions bit-for-bit across 36 replays per model on 12 inputs that vary observations, instructions, and noise. Compilation is not bit-exact: the maximum ten-step action differences are 0.010, 0.021, and 0.009 for S, B, and L. Our simulation success rates use uncompiled execution, which the CUDA graphs reproduce exactly, whereas the physical-robot La-JEPA policy runs compiled. For Fast-WAM and Faster-WAM, original-kernel graphs exceed the available GPU memory in our implementation.

\paragraph{Closed-loop validation.} Using the original-pipeline checkpoints (Appendix~\ref{app:libero-revision}), we evaluate the exact timed builds on 2{,}000 LIBERO episodes with a 700-step budget in every suite (Appendix~\ref{app:protocols}), with action noise of scale 1 redrawn at every replan, and compare each with uncompiled execution of the same checkpoint using paired McNemar tests. Compiled ten-step builds reach 95.20\%, 97.05\%, and 96.25\% for S, B, and L, against 94.30\%, 97.35\%, and 96.50\% uncompiled ($p=0.15$, $0.59$, and $0.69$). In separate evaluation runs, ten-step CUDA-graph builds reach 94.70\%, 97.00\%, and 96.35\% ($p=0.55$, $0.50$, and $0.84$), and the three-step B graph build 96.90\% against 97.30\% ($p=0.45$). None of the seven paired comparisons differs significantly. The timed L build applies LoRA to the final eight encoder blocks and is selected at 150k updates.

\paragraph{Serial encoding.} With serial rather than parallel text and vision encoding, optimized ten-step calls take 19, 26, and 61\,ms for S, B, and L, against 92\,ms for Fast-WAM and 54\,ms for Faster-WAM, leaving the speedups essentially unchanged.

\subsection{Parameter Accounting}
\label{app:param-accounting}

\paragraph{Definitions.} Table~\ref{tab:param-inventory} lists the components behind the size columns. Trainable counts the weights updated on target demonstrations, including LoRA adapters. Deployed counts the policy and encoder parameter inventory, including frozen visual and language encoders; it is a size measure, not a measurement of memory use or operations per action. Deployed counts merge LoRA adapters into the encoder weights, adding no parameters, and count shared encoders once across cameras. Baseline counts use measured checkpoints or published inventories; where encoder-only counts are unavailable, we count full tokenizer inventories, including unused decoder weights, which conservatively overestimates executed parameters. Transformer initialization refers to the trained world-action predictor or VLA backbone: video or image generation, vision-language pretraining, or random initialization.

\begin{figure}[t]
\centering
\includegraphics[width=\linewidth]{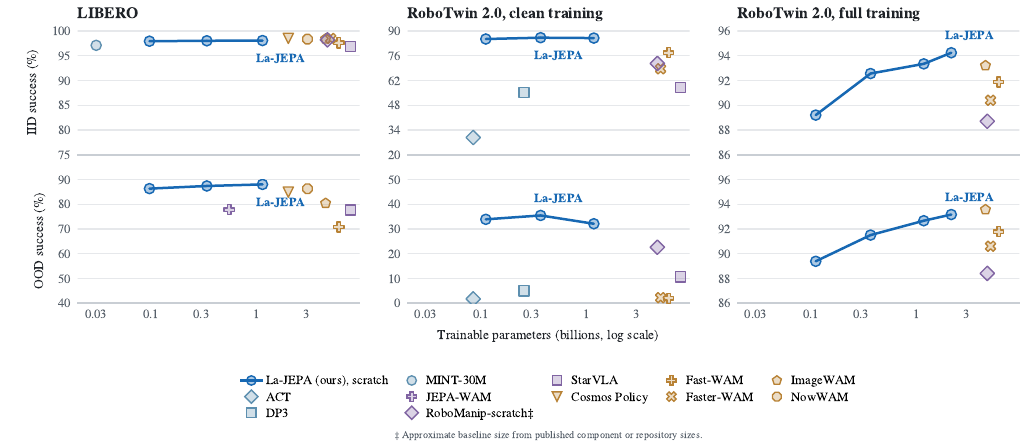}
\caption{Companion to Figure~\ref{fig:main-efficiency}: in-distribution (top) and out-of-distribution (bottom) success against trainable parameters on a logarithmic axis, for LIBERO (left), RoboTwin~2.0 with clean training (middle), and RoboTwin~2.0 with full clean and randomized training (right; La-JEPA XL from a single seed). Trainable counts every weight updated on the target demonstrations, adapters included. La-JEPA points are blue and connected in size order; purple denotes VLM-initialized and brown generator-initialized action transformers, while other scratch policies use muted blue. As in Figure~\ref{fig:main-efficiency}, LIBERO-Plus uses corrected instructions, and baselines without a corrected-instruction evaluation are omitted from that panel.}
\label{fig:main-efficiency-trainable}
\end{figure}

\paragraph{RoboTwin configurations.} Table~\ref{tab:param-inventory} refers to LIBERO configurations. RoboTwin's token projections change La-JEPA's sizes to 0.11/0.75\,B (trainable/deployed) for S, 0.37/1.01\,B for B, and 1.17/1.82\,B for L (Table~\ref{tab:robotwin-comparison}). Figure~\ref{fig:main-efficiency-trainable} complements Figure~\ref{fig:main-efficiency} by plotting success against trainable rather than deployed parameters.

\begin{table}[t]
\centering
\footnotesize
\renewcommand{\arraystretch}{1.1}
\caption{Parameter inventory (billions); La-JEPA rows for LIBERO. m: measured, r: reported, e: estimated.}
\label{tab:param-inventory}
\setlength{\tabcolsep}{4pt}
\input{tables/manuscript_param_inventory_table.tex}
\end{table}

\section{Real-Robot Setup}
\label{app:real_robot_setup}

This appendix describes the physical platform, the four tasks, demonstration collection, and the training and evaluation protocol behind Table~\ref{tab:real-robot}, and reports completion times.

\paragraph{Robot, cameras, and control.} A tabletop Franka Panda with a parallel gripper uses a wrist-mounted RealSense D405 camera and a fixed external RealSense D435 camera. A GPU workstation runs Linux with a real-time kernel. Policies receive both camera views, proprioception, and a task instruction, and execute end-effector deltas and gripper commands through Polymetis\footnote{\url{https://facebookresearch.github.io/fairo/polymetis/}} and the Franka Control Interface.

\paragraph{Tasks and objects.} Cube placement uses a red wooden cube and a bowl. Fruit picking requires removing a bamboo steamer's lid, setting it down, and then picking the correct fruit, a plum, from inside the steamer. Push T requires pushing a wooden T onto a T shape marked on the table. Dish manipulation requires placing both plates, then the spoon on the gray plate and the fork on the pink plate, in one uninterrupted rollout; plate locations are fixed, but the utensils' starting positions are interchangeable. Every training demonstration and evaluation rollout contains the complete uninterrupted sequence of its task. Figure~\ref{fig:real-robot-tasks} shows the initial scene of each task from the external camera.

\paragraph{Demonstration collection.} SpaceMouse teleoperation provides six-degree-of-freedom end-effector control through the same interface. We record camera observations, robot state, end-effector pose, and gripper state at 10\,Hz, collecting 100 demonstrations per task (400 in total). Initial cube positions vary within $10\,\mathrm{cm}\times10\,\mathrm{cm}$, and plum positions vary within the steamer; Push T and dish manipulation use smaller placement variations, including utensil rearrangements.

\paragraph{Training and evaluation.} Policies train jointly on the four language-conditioned tasks using their respective recipes. Fast-WAM and Faster-WAM train for 30 epochs; La-JEPA trains its B policy transformer from scratch with encoder LoRA but without image augmentation. Table~\ref{tab:real-robot} reports counts over 20 rollouts per task, with explicit ten-rollout denominators for LingBot-VA's dish result. Utensil~1 and Utensil~2 denote successful placement of the first and second utensils on plates, irrespective of utensil identity or order. Position additionally requires the correct spoon--gray and fork--pink assignments and success at all preceding stages; the two plate stages share one table column because their counts are identical. Average weights the four tasks equally, using Position for dish, in mean successes out of 20; LingBot-VA's average is omitted because its dish result uses 10 trials rather than 20.

\paragraph{Completion time.} Table~\ref{tab:robot-times} reports the mean and standard deviation of completion time over successful trials. La-JEPA is the fastest policy without embodied pretraining on every task and the fastest overall on fruit picking and Push T, while $\pi_{0.5}$ is fastest on cube placement and dish manipulation. Our main hypothesis for LingBot-VA's low Push T and dish success is that it requires substantially more real-robot finetuning than the other policies; independently of success, its completion times, three to eleven times longer than those of the other policies, make it impractical for closed-loop control on this platform. Its dish entry comes from its single successful trial.

\begin{table}[h]
\centering\small
\caption{Real-robot completion time over successful trials (s; standard deviation in small type). Bold: fastest; underlined: second. $^{\dagger}$Single successful trial.}
\label{tab:robot-times}
\setlength{\tabcolsep}{4pt}
\input{tables/manuscript_robot_completion_times_table.tex}
\end{table}

\section{Supplementary Results}
\label{app:supplementary}
\begin{table}[p]
\centering
\small
\renewcommand{\arraystretch}{0.98}
\caption{Per-task RoboTwin~2.0 success of La-JEPA S, B, and L (\%), mean over three training seeds with 100 episodes per task, condition, and seed.}
\label{tab:per-task-robotwin}
\setlength{\tabcolsep}{6pt}
\input{tables/per_task_robotwin_table.tex}
\end{table}

This appendix collects results that inform but do not define the final recipe: adapted future targets, evaluation-time settings, future-only pooling, representation diagnostics, and per-task RoboTwin success.

\subsection{Adapting Future Targets}
\label{app:adapted-targets}

On LIBERO with L, we replace the frozen future targets with targets from an EMA of the adapting encoder, keeping the clips, observation processing, and normalization statistics of the frozen-target policy and passing no gradients through the target path. The adapted-target policy, selected at 90k updates, achieves 81.65\% on LIBERO-Plus, against 82.26\% for the frozen-target policy at 120k, both trained with the original pipeline (Appendix~\ref{app:libero-revision}) and at action-noise scale 1. Adapted targets do not improve success (paired McNemar $p=0.096$), so the final recipe retains frozen targets.

\subsection{Evaluation Settings}
\label{app:eval-settings}

At the 120k checkpoints of the original-pipeline policies (Appendix~\ref{app:libero-revision}), we vary the execution chunk length, the number of denoising steps, and the action-noise scale for L and B, and test three-step inference for S (Table~\ref{tab:eval-sweeps}); the reference executes ten actions per replan with ten denoising steps and noise scale 1, and LIBERO runs use a 700-step budget in every suite (Appendix~\ref{app:protocols}). Executing only five actions reduces LIBERO-Plus success by 3.29 points for L and 3.45 for B, whereas the number of denoising steps matters less: with three steps, B and L stay within one point of their ten-step references on both benchmarks, and S obtains 94.50\% against 94.30\% on LIBERO ($p=0.81$, paired McNemar). Three-step calls take 23\,ms for B and 39\,ms for L with original-kernel CUDA graphs (Table~\ref{tab:latency}), an operating point separate from the ten-step latency comparison. No tested noise scale gives a statistically supported aggregate improvement in the paired comparisons; the original L policy reaches 82.26\% at the default scale 1 (Appendix~\ref{app:action-noise}), and the best single scale, 0.5, reaches 82.50\%.

\begin{table}[t]
\centering
\caption{Evaluation-setting sweeps at fixed checkpoints (success in \%). LIBERO values use the 700-step budget in every suite (Appendix~\ref{app:protocols}); B's LIBERO reference is a re-evaluation of its default setting.}
\label{tab:eval-sweeps}
\begin{minipage}[t]{0.49\linewidth}
\centering\footnotesize
\textbf{LIBERO}\par\smallskip
\setlength{\tabcolsep}{3pt}
\input{tables/eval_sweeps_iid_table.tex}
\end{minipage}\hfill
\begin{minipage}[t]{0.49\linewidth}
\centering\footnotesize
\textbf{LIBERO-Plus}\par\smallskip
\setlength{\tabcolsep}{3pt}
\input{tables/eval_sweeps_ood_table.tex}
\end{minipage}
\end{table}

\subsection{Future-Only Pooling}
\label{app:future-pooling}

Complementing the token-reduction comparison (Appendix~\ref{app:token-reduction}), we keep patch-grouped observations and mean-pool only the future targets (Table~\ref{tab:future-only-pooling}). In-distribution success stays near 61--62\%, whereas out-of-distribution success is non-monotonic in pooling granularity. Because the settings also differ in positional geometry, target normalization, camera weighting, and checkpoint step, these scores describe the tested recipes rather than an isolated effect of target resolution. The policies are small, frozen-encoder, static-feature policies trained on the 95\% split. Observations keep patch grouping with spatial factors $[2,4,4]$ for the head, left-wrist, and right-wrist cameras, and the future targets are pooled by $[2,4,4]$, $[4,8,8]$, or global averaging, giving $[80,20,20]$, $[20,6,6]$, or $[1,1,1]$ spatial tokens per temporal position. The first setting has the patch-grouped reference's token count but averages rather than concatenates features. The rows report 150k, 60k, and 90k checkpoints, each evaluated in both regimes on 5{,}000 episodes, under an evaluation implementation that differs from the other studies.

\begin{table}[t]
\centering
\caption{Future-only pooling on RoboTwin (success in \%).}
\label{tab:future-only-pooling}
\input{tables/manuscript_future_pooling_table.tex}
\end{table}

\subsection{Representation Diagnostics}
\label{app:representations}

Table~\ref{tab:historical-objective-diagnostics} analyzes frozen-encoder RoboTwin policies trained with and without future prediction at 150k updates, separate from the policies in Table~\ref{tab:design-additive} (block A). Centered RankMe~\citep{rankme} uses 2{,}000 sampled, mean-pooled final-layer value features, and a cosine 5-nearest-neighbor action readout reports $R^2$ over normalized action chunks. Features span 100 episodes held out from policy training, with an episode-disjoint 70/30 split for fitting and testing the readout and with language and proprioceptive conditioning retained. Future prediction raises effective rank together with clean success under the 1{,}000-episode protocol, but lowers action-readout $R^2$, so nearest-neighbor action decodability does not explain the difference in control performance.

\begin{table}[t]
\centering
\caption{Representation diagnostics with and without future prediction on RoboTwin.}
\label{tab:historical-objective-diagnostics}
\input{tables/manuscript_representation_diagnostics_table.tex}
\end{table}

\subsection{Per-Task RoboTwin Results}
\label{app:per-task}

Table~\ref{tab:per-task-robotwin} lists the per-task RoboTwin~2.0 success of La-JEPA S, B, and L, the policies reported in Table~\ref{tab:robotwin-comparison}, trained on clean demonstrations and evaluated on 100 episodes per task in clean and randomized scenes. Success varies widely across tasks, and randomized scenes lower success on almost every task.

\end{document}

%% file: math_commands.tex
\usepackage{amsmath,amsfonts,bm}

\def\eqref#1{equation~\ref{#1}}

\def\1{\bm{1}}

\DeclareMathAlphabet{\mathsfit}{\encodingdefault}{\sfdefault}{m}{sl}
\SetMathAlphabet{\mathsfit}{bold}{\encodingdefault}{\sfdefault}{bx}{n}

%% file: tables/manuscript_main_comparison_table.tex
\begin{tabular}{llrrccc}
\toprule
 & & \multicolumn{2}{c}{Params (B)} & & \multicolumn{2}{c}{LIBERO-Plus} \\
\cmidrule(lr){3-4}\cmidrule(lr){6-7}
Method & Transformer init. & Train. & Depl. & LIBERO & Corrected & Released \\
\midrule
Qwen-RobotManip-scratch & VLM (supervised) & 4.7 & 4.7 & 98.2 & --- & 78.3 \\
StarVLA & VLM (supervised) & 7.8 & 7.8 & 96.9 & 77.6 & 76.7 \\
JEPA-WAM & VLM (supervised) & 0.55 & 1.35 & 96.7 & 77.8 & 77.1 \\
Fast-WAM & video generator & 6.0 & 12.4 & 97.6 & 70.8 & 51.5 \\
Faster-WAM & video generator & 5.1 & 11.4 & \textbf{98.5} & --- & 75.1 \\
Cosmos Policy & video generator & 2.0 & 7.0 & \textbf{98.5} & 84.9 & 80.8 \\
ImageWAM & image generator & 4.5 & 8.6 & 98.4 & 80.4 & 79.7 \\
NowWAM$^{\S}$ & image generator & 3.0 & 7.1 & 98.4 & 86.2 & \textbf{86.2} \\
\midrule
La-JEPA S (ours) & scratch & 0.10 & 0.74 & 97.9 & 86.3 & 80.0 \\
La-JEPA B (ours) & scratch & 0.34 & 0.98 & 98.0 & \underline{87.3} & 82.5 \\
La-JEPA L (ours) & scratch & 1.14 & 1.78 & 98.1 & \textbf{88.0} & \underline{84.1} \\
\bottomrule
\end{tabular}

%% file: tables/manuscript_robotwin_comparison_table.tex
\begin{tabular}{llrrccc}
\toprule
 & & \multicolumn{2}{c}{Params (B)} & \multicolumn{3}{c}{RoboTwin~2.0 success (\%)} \\
\cmidrule(lr){3-4}\cmidrule(lr){5-7}
Method & Transformer init. & Train. & Depl. & Clean & Randomized & Mean \\
\midrule
\multicolumn{7}{l}{\emph{Clean training: clean demonstrations only}} \\
Qwen-RobotManip-scratch & VLM (supervised) & 4.7 & 4.7 & 71.6 & 22.6 & 47.1 \\
StarVLA & VLM (supervised) & 7.8 & 7.8 & 58.1 & 10.6 & 34.4 \\
Fast-WAM & video generator & 6.0 & 12.4 & 77.8 & 1.9 & 39.9 \\
Faster-WAM & video generator & 5.1 & 11.4 & 68.6 & 2.2 & 35.4 \\
\midrule
La-JEPA S (ours) & scratch & 0.11 & 0.75 & 85.5 & \underline{33.9} & \underline{59.7} \\
La-JEPA B (ours) & scratch & 0.37 & 1.01 & \textbf{86.2} & \textbf{35.4} & \textbf{60.8} \\
La-JEPA L (ours) & scratch & 1.17 & 1.82 & \underline{86.0} & 32.1 & 59.0 \\
\midrule
\multicolumn{7}{l}{\emph{Full training: clean and randomized demonstrations}} \\
Qwen-RobotManip-scratch & VLM (supervised) & 4.7 & 4.7 & 88.7 & 88.4 & 88.6 \\
Fast-WAM & video generator & 6.0 & 12.4 & 91.9 & 91.8 & 91.8 \\
Faster-WAM & video generator & 5.1 & 11.4 & 90.4 & 90.6 & 90.5 \\
ImageWAM & image generator & 4.5 & 8.6 & 93.2 & \textbf{93.6} & \underline{93.4} \\
\midrule
La-JEPA S (ours) & scratch & 0.11 & 0.75 & 89.2 & 89.4 & 89.3 \\
La-JEPA B (ours) & scratch & 0.37 & 1.01 & 92.6 & 91.5 & 92.0 \\
La-JEPA L (ours) & scratch & 1.17 & 1.82 & \underline{93.3} & 92.7 & 93.0 \\
La-JEPA XL (ours) & scratch & 2.14 & 2.79 & \textbf{94.2} & \underline{93.2} & \textbf{93.7} \\
\bottomrule
\end{tabular}

%% file: tables/manuscript_jepa_comparison_table.tex
\begin{tabular}{llcrrrr}
\toprule
 & & & \multicolumn{2}{c}{Params (B)} & & \\
\cmidrule(lr){4-5}
Method & Transformer init. & Embodied PT & Train. & Depl. & LIBERO & Plus \\
\midrule
LDA-1B$^{\ddagger}$ & --- & yes & --- & --- & 90.6 & 45.5 \\
LeapBot-WA & video generator & yes & 2.7 & 10.4 & 97.3 & 71.7 \\
JEPA-WAM & VLM (supervised) & no & 0.55 & 1.35 & 96.7 & 77.1 \\
La-JEPA S (ours) & scratch & no & 0.10 & 0.74 & 97.9 & 80.0 \\
La-JEPA B (ours) & scratch & no & 0.34 & 0.98 & \underline{98.0} & \underline{82.5} \\
La-JEPA L (ours) & scratch & no & 1.14 & 1.78 & \textbf{98.1} & \textbf{84.1} \\
\midrule
 & Training tasks & & & & C2C & C2R \\
\midrule
JEPA-WAM & 20 & no & 0.55 & 1.35 & 79.9 & 36.9 \\
La-JEPA B (ours) & 20 & no & 0.37 & 1.01 & \textbf{95.2} & \textbf{56.5} \\
\bottomrule
\end{tabular}

%% file: tables/manuscript_real_robot_table.tex
\begin{tabular}{lrrrrrrrrr}
\toprule
\raisebox{-2ex}[0pt][0pt]{Policy} & \raisebox{-2ex}[0pt][0pt]{Latency (ms)} & \raisebox{-2ex}[0pt][0pt]{Fruit} & \raisebox{-2ex}[0pt][0pt]{Cube} & \raisebox{-2ex}[0pt][0pt]{Push T} & \multicolumn{4}{c}{Dish manipulation} & \raisebox{-2ex}[0pt][0pt]{Average} \\
\cmidrule(lr){6-9}
 &  &  &  &  & Plates & Utensil 1 & Utensil 2 & Position &  \\
\midrule
\multicolumn{10}{l}{\textit{With embodied pretraining}} \\
$\pi_{0.5}$ & 71.89 & \textbf{20} & \textbf{20} & 11 & \textbf{20} & \textbf{19} & \textbf{19} & \textbf{18} & 17.25 \\
LingBot-VA & $\approx 4760$ & 18 & \textbf{20} & 2 & 6/10 & 3/10 & 2/10 & 1/10 & --- \\
\midrule
\multicolumn{10}{l}{\textit{Without embodied pretraining}} \\
Fast-WAM & 88.09 & 19 & 19 & 14 & \textbf{20} & \textbf{19} & 15 & 12 & 16.00 \\
Faster-WAM & 50.96 & 18 & \textbf{20} & 13 & \textbf{20} & 18 & 16 & 14 & 16.25 \\
La-JEPA & 25.16 & 19 & \textbf{20} & \textbf{15} & \textbf{20} & 18 & 17 & 16 & \textbf{17.50} \\
\bottomrule
\end{tabular}

%% file: tables/ablation_loo_table.tex
\begin{tabular}{lllll}
\toprule
 & \multicolumn{2}{c}{RoboTwin~2.0} & & \\
\cmidrule(lr){2-3}
Removed from the final recipe & Clean & Randomized & LIBERO & LIBERO-Plus \\
\midrule
None (reference S) & 85.6 & 33.1 & 97.9 & 79.5 \\
\midrule
Future prediction (action only) & 80.8 ($-$4.8 [$-$5.9, $-$3.7]) & 30.2 ($-$2.9 [$-$4.2, $-$1.5]) & 95.7 ($-$2.2 [$-$3.3, $-$1.2]) & 75.0 ($-$4.5 [$-$5.4, $-$3.6]) \\
V-JEPA features (Wan2.2 VAE) & 72.0 ($-$13.7) & 4.7 ($-$28.4) & 96.5 ($-$1.4 [$-$2.4, $-$0.4]) & 68.9 ($-$10.7 [$-$11.6, $-$9.7]) \\
Tube-motion (tube) & 85.7 ($+$0.0) & 34.2 ($+$1.1) & 97.5 ($-$0.4, ns) & 78.5 ($-$1.1 [$-$1.8, $-$0.3]) \\
Tube-motion (static) & 86.4 ($+$0.8) & 32.9 ($-$0.2) & --- & --- \\
Camera dropout & 85.0 ($-$0.7, ns) & 35.9 ($+$2.8 [$+$1.5, $+$4.1])$^{\dagger}$ & 97.8 ($-$0.2, ns) & 75.5 ($-$4.0 [$-$4.8, $-$3.2]) \\
Proprioception dropout & 84.8 ($-$0.8, ns) & 31.9 ($-$1.2, ns) & 98.4 ($+$0.5, ns) & 78.6 ($-$1.0 [$-$1.7, $-$0.2]) \\
Image augmentation & 83.2 ($-$2.4 [$-$3.4, $-$1.3]) & 18.0 ($-$15.1 [$-$16.4, $-$13.8]) & 95.6 ($-$2.4) & 66.5 ($-$13.0 [$-$14.0, $-$12.1]) \\
Encoder LoRA & 81.7 ($-$4.0) & 32.1 ($-$1.0) & 97.1 ($-$0.9) & 74.8 ($-$4.7) \\
\midrule
All three robustness components & 82.1 ($-$3.5 [$-$4.6, $-$2.4]) & 18.5 ($-$14.6 [$-$15.9, $-$13.3]) & 96.6 ($-$1.3) & 67.3 ($-$12.2 [$-$13.2, $-$11.2]) \\
V-JEPA features and all robustness & 60.5 ($-$25.1) & 0.7 ($-$32.4) & 88.6 ($-$9.4 [$-$10.8, $-$7.9]) & 36.1 ($-$43.4 [$-$44.5, $-$42.3]) \\
\bottomrule
\end{tabular}

%% file: tables/ablation_deltas_table.tex
\begin{tabular}{llllrrl}
\toprule
 & Choice & Baseline & Benchmark & IID & OOD & Evaluation \\
\midrule
\multicolumn{7}{l}{\textbf{A}\quad Future prediction} \\
 & Action + future prediction & Action only & LIBERO & $+$0.9 & $+$2.1 & IID: seed 42; OOD: 3 seeds \\
 & Action + future prediction & Action only & RoboTwin & $+$7.0 & $+$3.7 & $5{,}000$/cond., 120--150k \\
\midrule
\multicolumn{7}{l}{\textbf{B}\quad V-JEPA features} \\
 & V-JEPA~2.1 features & Wan2.2 VAE latents & LIBERO & $+$2.0 & $+$6.0 & $2{,}000$ / $10{,}030$, 150k \\
 & V-JEPA~2.1 features & Wan2.2 VAE latents & RoboTwin & $+$15.9 & $+$16.0 & $5{,}000$/cond., 150k \\
\midrule
\multicolumn{7}{l}{\textbf{C}\quad V-JEPA setup} \\
 & Tube & Static & RoboTwin & $+$1.1 & $+$3.6 & $1{,}000$/cond., 120--150k \\
 & Tube-motion & Static & RoboTwin & $+$3.4 & $+$3.7 & $1{,}000$/cond., 120--150k \\
\midrule
\multicolumn{7}{l}{\textbf{D}\quad Camera dropout} \\
 & Camera dropout & Wrist-only camera dropout & LIBERO & $+$0.4 & $+$10.2 & $n{=}2{,}000$ / $10{,}030$, 90k \\
 & Camera dropout & No camera dropout & RoboTwin & $-$0.6 & $+$7.1 & $1{,}000$/cond., 150k \\
\midrule
\multicolumn{7}{l}{\textbf{E}\quad Proprioception dropout} \\
 & Proprioception dropout & No proprioception dropout & RoboTwin & $+$3.8 & $+$4.2 & $1{,}000$/cond., best step \\
\midrule
\multicolumn{7}{l}{\textbf{F}\quad Augmentation and encoder LoRA} \\
 & Encoder LoRA & Plain & LIBERO & $+$0.5 & $+$1.4 & $n{=}2{,}000$ / $10{,}030$ \\
 & Image augmentation & Plain & LIBERO & $+$0.9 & $+$5.8 & $n{=}2{,}000$ / $10{,}030$ \\
 & Augmentation + LoRA & Plain & LIBERO & $+$0.1 & $+$7.2 & $n{=}2{,}000$ / $10{,}030$ \\
 & Augmentation + LoRA & Plain & RoboTwin & $+$5.7 & $+$16.0 & $5{,}000$/cond., 150k \\
\bottomrule
\end{tabular}

%% file: tables/objective_seed_comparison_table.tex
\begin{tabular}{rrrrrr}
\toprule
Seed & Selected step & Joint & Action only & Gain (pp) & Paired $p$ \\
\midrule
42 & 150k & 76.13 & 74.47 & $+$1.67 & $2.3\times10^{-4}$ \\
1 & 90k & 76.13 & 74.77 & $+$1.37 & $0.002$ \\
2 & 120k & 77.15 & 73.82 & $+$3.33 & $2.2\times10^{-12}$ \\
\midrule
Mean & --- & 76.47 & 74.35 & $+$2.12 & --- \\
\bottomrule
\end{tabular}

%% file: tables/manuscript_libero_perturbations_table.tex
\begin{tabular}{lrr}
\hline
Training variant & Camera viewpoints & Sensor noise \\
\hline
\multicolumn{3}{l}{Frozen encoder, stored features} \\
Wrist-camera dropout only & 11.4 & 42.0 \\
Either-camera dropout & 43.8 & 67.1 \\
\hline
\multicolumn{3}{l}{Online encoder, either-camera dropout} \\
Image augmentation only & 74.6 & 81.4 \\
Encoder LoRA only & 43.9 & 73.3 \\
Image augmentation + encoder LoRA & 72.5 & 83.6 \\
\hline
\end{tabular}

%% file: tables/manuscript_protocol_settings_table.tex
\begin{tabular}{@{}p{0.25\linewidth}lll@{}}
\toprule
Policy size & S & B & L \\
Width / layers / heads & 384 / 12 / 6 & 768 / 12 / 12 & 1024 / 24 / 16 \\
LoRA blocks (LIBERO) & Last 8 & Last 8 & All 24 \\
LoRA blocks (RoboTwin) & Last 8 & Last 8 & Last 8 \\
\midrule
Transformer / losses & \multicolumn{3}{p{0.69\linewidth}}{Equal World/Action DiT widths and depths; MLP expansion 4; action and future loss weights 1 each} \\
Optimizer & \multicolumn{3}{p{0.69\linewidth}}{AdamW; policy LR $5\times10^{-4}$; betas $(0.9,0.95)$; matrix weight decay 0.01; gradient clipping 1} \\
Precision / EMA & \multicolumn{3}{p{0.69\linewidth}}{BF16 mixed precision; EMA decay 0.9999} \\
Schedule & \multicolumn{3}{p{0.69\linewidth}}{150{,}000 updates; 5\% warmup, cosine decay, final 10\% constant tail; LR floor 1\% of peak} \\
Global batch & \multicolumn{3}{p{0.69\linewidth}}{LIBERO: 128; RoboTwin: 400} \\
LoRA & \multicolumn{3}{p{0.69\linewidth}}{Rank 16; $\alpha=32$; dropout 0.05; LR $10^{-4}$ with cosine decay to 5\% of peak and 1\% in the final tail; weight decay 0} \\
Adapter locations & \multicolumn{3}{p{0.69\linewidth}}{Fused Q/K/V, attention output, both MLP projections} \\
Conditioning dropout & \multicolumn{3}{p{0.69\linewidth}}{Text and proprioception: 0.1 each; camera: 0.1, either view on LIBERO, wrists only on RoboTwin; retain at least one view} \\
Action/state normalization & \multicolumn{3}{p{0.69\linewidth}}{Per-channel 1st/99th percentiles mapped to $[-1,1]$; clamped to $\pm5$ on RoboTwin} \\
Feature normalization & \multicolumn{3}{p{0.69\linewidth}}{Training-set statistics; separate motion-target statistics; LIBERO pooled futures use per-camera scalar variance correction from 64 training windows} \\
Inference defaults & \multicolumn{3}{p{0.69\linewidth}}{$\tau_{\mathrm{ctx}}=0$; predict 32 actions; execute 10 per replan on LIBERO, 24 on RoboTwin} \\
\bottomrule
\end{tabular}

%% file: tables/manuscript_protocol_augmentation_table.tex
\begin{tabular}{@{}p{0.18\linewidth}p{0.19\linewidth}p{0.55\linewidth}@{}}
\toprule
Operation & Probability & Parameters and scope \\
\midrule
Unmodified window & 0.3 & Bypass all observation augmentation \\
Photometric & 0.6 & All cameras share brightness/contrast/saturation strength 0.3, hue 0.03 turns, gamma $[0.8,1.25]$ \\
Corruption & 0.5 & Primary camera; uniform choice of Gaussian blur ($\sigma\in[0.5,6]$ pixels), motion blur (length $[5,20]$ pixels), or fog \\
Fog & Within corruption & Opacity $[0.2,0.6]$; contrast multiplier $[0.5,0.9]$; grid size 4--8; gray fraction 0.5--1.0 \\
Rotation & 0.5 & Primary camera; yaw/pitch $[-4,4]^\circ$, roll $[-3,3]^\circ$; homography $KRK^{-1}$, replicated borders; vertical FOV $45^\circ$ (LIBERO), $37^\circ$ (RoboTwin) \\
\bottomrule
\end{tabular}

%% file: tables/manuscript_latency_table.tex
\begin{tabular}{lrrrr}
\toprule
Policy & Steps & Optimized & \shortstack{Original-kernel\\CUDA graphs} & Eager \\
\midrule
La-JEPA S & 10 & 17.78 & 34.20 & 99.96 \\
La-JEPA B & 10 & 25.16 & 42.35 & 98.21 \\
La-JEPA L & 10 & 59.72 & 88.87 & 167.93 \\
Fast-WAM & 10 & 88.09 & OOM & 193.44 \\
Faster-WAM & 10 & 50.96 & OOM & 62.70 \\
$\pi_{0.5}$ & 10 & 71.89 & --- & --- \\
\midrule
La-JEPA S & 3 & 11.04 & 19.59 & --- \\
La-JEPA B & 3 & 13.40 & 22.54 & --- \\
La-JEPA L & 3 & 26.30 & 39.37 & --- \\
\bottomrule
\end{tabular}

%% file: tables/manuscript_param_inventory_table.tex
\begin{tabular}{l>{\raggedright\arraybackslash}m{3.4cm}>{\raggedright\arraybackslash}m{3.2cm}rr}
\toprule
Method & Trained components & Frozen components at inference & Trainable & Deployed \\
\midrule
\addlinespace[2pt]
Qwen-RobotManip-scratch & Qwen3.5-4B, action head & --- & 4.7$^{e}$ & 4.7$^{e}$ \\
\addlinespace[4pt]
StarVLA & Qwen3-VL-4B (4.44), cross-attention DiT action head (3.36) & --- & 7.79$^{m}$ & 7.79$^{m}$ \\
\addlinespace[4pt]
Fast-WAM & Wan2.2 video DiT (5.00), action DiT (1.02) & UMT5-XXL encoder (5.68), Wan2.2 VAE (0.71) & 6.02$^{m}$ & 12.4$^{m}$ \\
\addlinespace[4pt]
Faster-WAM & Wan2.2 video DiT (5.00), action head (0.05) & UMT5-XXL encoder (5.68), Wan2.2 VAE (0.71) & 5.05$^{r}$ & 11.4$^{e}$ \\
\addlinespace[4pt]
Cosmos Policy & Cosmos-Predict2-2B DiT (1.96) & T5-11B encoder (4.87), Wan2.1 VAE (0.13) & 1.96$^{m}$ & 6.95$^{m}$ \\
\addlinespace[4pt]
ImageWAM & FLUX.2 4B image DiT (3.86), action DiT (0.64) & Qwen3-4B encoder (4.02), VAE (0.08) & 4.52$^{m}$ & 8.62$^{m}$ \\
\addlinespace[4pt]
NowWAM & FLUX.2-klein 4B-initialized DiT (3.04) & Qwen3-4B encoder (4.02), autoencoder (0.08) & 3.04$^{m}$ & 7.1$^{m}$ \\
\addlinespace[4pt]
JEPA-WAM & action DiT (0.53), Qwen2.5-0.5B LoRA (0.018), transition head (0.004, training only) & Qwen2.5-0.5B (0.49), V-JEPA~2.1 L (0.30), projector (0.002) & 0.55$^{m}$ & 1.35$^{m}$ \\
\addlinespace[4pt]
LeapBot-WA & mixture-of-transformers policy (2.75) & V-JEPA~2.1 encoder (1.85), adapter (0.11), UMT5-XXL encoder (5.68) & 2.75$^{m}$ & 10.4$^{m}$ \\
\addlinespace[4pt]
$\pi_{0.5}$ & PaliGemma (2.92), action expert (0.43) & --- & 3.35$^{m}$ & 3.35$^{m}$ \\
\addlinespace[2pt]
\midrule
\addlinespace[2pt]
La-JEPA S & world and action transformers (0.094), encoder LoRA (0.002) & V-JEPA~2.1 L (0.305), FLAN-T5-large encoder (0.341) & 0.096$^{m}$ & 0.74$^{m}$ \\
\addlinespace[4pt]
La-JEPA B & world and action transformers (0.335), encoder LoRA (0.002) & V-JEPA~2.1 L (0.305), FLAN-T5-large encoder (0.341) & 0.337$^{m}$ & 0.98$^{m}$ \\
\addlinespace[4pt]
La-JEPA L & world and action transformers (1.131), full-depth encoder LoRA (0.006) & V-JEPA~2.1 L (0.305), FLAN-T5-large encoder (0.341) & 1.14$^{m}$ & 1.78$^{m}$ \\
\addlinespace[2pt]
\bottomrule
\end{tabular}

%% file: tables/manuscript_robot_completion_times_table.tex
\begin{tabular}{lrrrrr}
\toprule
 & \multicolumn{2}{c}{With embodied pretraining} & \multicolumn{3}{c}{Without embodied pretraining} \\
\cmidrule(lr){2-3}\cmidrule(lr){4-6}
Task & $\pi_{0.5}$ & LingBot-VA & Fast-WAM & Faster-WAM & La-JEPA \\
\midrule
Fruit & \underline{24.4}{\scriptsize\,$\pm$\,2.7} & 175.3{\scriptsize\,$\pm$\,41.0} & 36.7{\scriptsize\,$\pm$\,9.0} & 35.1{\scriptsize\,$\pm$\,9.5} & \textbf{21.4}{\scriptsize\,$\pm$\,0.8} \\
Cube & \textbf{10.8}{\scriptsize\,$\pm$\,0.5} & 117.8{\scriptsize\,$\pm$\,50.9} & 21.9{\scriptsize\,$\pm$\,7.6} & 18.2{\scriptsize\,$\pm$\,4.6} & \underline{11.5}{\scriptsize\,$\pm$\,3.2} \\
Push T & \underline{22.8}{\scriptsize\,$\pm$\,14.7} & 112.5{\scriptsize\,$\pm$\,9.1} & 24.6{\scriptsize\,$\pm$\,6.6} & 23.8{\scriptsize\,$\pm$\,12.1} & \textbf{21.1}{\scriptsize\,$\pm$\,4.1} \\
Dish & \textbf{39.0}{\scriptsize\,$\pm$\,3.4} & 269.0$^{\dagger}$ & 85.0{\scriptsize\,$\pm$\,15.3} & 72.0{\scriptsize\,$\pm$\,8.8} & \underline{58.1}{\scriptsize\,$\pm$\,8.7} \\
\bottomrule
\end{tabular}

%% file: tables/per_task_robotwin_table.tex
\begin{tabular}{lrrrrrr}
\toprule
 & \multicolumn{2}{c}{S} & \multicolumn{2}{c}{B} & \multicolumn{2}{c}{L} \\
\cmidrule(lr){2-3}\cmidrule(lr){4-5}\cmidrule(lr){6-7}
Task & Clean & Rand. & Clean & Rand. & Clean & Rand. \\
\midrule
Adjust bottle & 100.0 & 21.0 & 100.0 & 18.7 & 99.7 & 15.7 \\
Beat block hammer & 91.3 & 17.7 & 94.7 & 20.7 & 94.0 & 21.0 \\
Blocks ranking rgb & 95.7 & 35.7 & 95.0 & 43.7 & 97.0 & 43.3 \\
Blocks ranking size & 67.3 & 29.3 & 69.3 & 34.0 & 66.7 & 29.7 \\
Click alarmclock & 99.7 & 47.7 & 100.0 & 68.0 & 99.7 & 64.3 \\
Click bell & 100.0 & 63.7 & 100.0 & 64.3 & 100.0 & 64.0 \\
Dump bin bigbin & 93.0 & 79.7 & 94.7 & 81.7 & 94.3 & 79.0 \\
Grab roller & 100.0 & 45.7 & 100.0 & 46.3 & 99.7 & 44.3 \\
Handover block & 88.3 & 5.0 & 93.0 & 7.0 & 86.3 & 5.7 \\
Handover mic & 99.7 & 8.3 & 99.3 & 4.0 & 98.3 & 6.7 \\
Hanging mug & 47.7 & 3.7 & 49.3 & 7.3 & 54.0 & 5.0 \\
Lift pot & 99.7 & 0.0 & 99.7 & 0.3 & 97.3 & 0.7 \\
Move can pot & 84.7 & 4.3 & 83.3 & 4.7 & 81.3 & 4.0 \\
Move pillbottle pad & 90.0 & 10.3 & 94.3 & 10.3 & 91.0 & 10.3 \\
Move playingcard away & 96.0 & 8.0 & 96.3 & 21.0 & 96.7 & 23.0 \\
Move stapler pad & 50.0 & 4.7 & 48.3 & 5.0 & 32.3 & 2.3 \\
Open laptop & 95.0 & 6.3 & 95.7 & 6.7 & 96.0 & 5.7 \\
Open microwave & 91.0 & 14.0 & 87.7 & 13.0 & 89.3 & 16.7 \\
Pick diverse bottles & 80.3 & 30.7 & 84.7 & 39.7 & 82.3 & 37.7 \\
Pick dual bottles & 92.7 & 35.0 & 92.0 & 44.7 & 91.7 & 43.3 \\
Place a2b left & 68.3 & 32.3 & 74.3 & 27.7 & 81.3 & 24.0 \\
Place a2b right & 67.3 & 26.0 & 66.7 & 19.7 & 71.3 & 19.3 \\
Place bread basket & 88.0 & 56.3 & 90.0 & 58.0 & 93.3 & 53.7 \\
Place bread skillet & 91.7 & 30.3 & 93.3 & 29.3 & 91.0 & 31.0 \\
Place burger fries & 97.7 & 82.7 & 98.7 & 74.0 & 93.3 & 55.0 \\
Place can basket & 74.7 & 16.3 & 73.0 & 17.3 & 81.0 & 20.7 \\
Place cans plasticbox & 99.3 & 60.0 & 99.3 & 63.7 & 99.3 & 53.3 \\
Place container plate & 100.0 & 71.7 & 98.7 & 78.0 & 98.7 & 72.7 \\
Place dual shoes & 69.3 & 39.3 & 71.0 & 32.3 & 78.0 & 27.3 \\
Place empty cup & 100.0 & 56.7 & 100.0 & 50.7 & 100.0 & 55.3 \\
Place fan & 68.0 & 10.3 & 71.7 & 5.7 & 76.0 & 6.3 \\
Place mouse pad & 65.0 & 7.7 & 71.0 & 7.0 & 71.0 & 5.0 \\
Place object basket & 85.0 & 32.0 & 83.7 & 32.7 & 83.7 & 28.3 \\
Place object scale & 76.3 & 15.0 & 82.3 & 23.0 & 81.7 & 18.0 \\
Place object stand & 96.0 & 49.7 & 96.3 & 60.7 & 97.0 & 51.3 \\
Place phone stand & 85.7 & 26.0 & 83.3 & 30.0 & 77.7 & 21.0 \\
Place shoe & 94.3 & 57.3 & 92.0 & 54.3 & 95.3 & 39.3 \\
Press stapler & 95.7 & 68.3 & 92.3 & 58.7 & 90.7 & 44.3 \\
Put bottles dustbin & 82.0 & 72.7 & 81.0 & 78.0 & 69.7 & 60.3 \\
Put object cabinet & 37.7 & 37.0 & 39.7 & 26.7 & 41.3 & 22.3 \\
Rotate qrcode & 89.0 & 4.7 & 88.7 & 6.0 & 88.0 & 4.0 \\
Scan object & 84.0 & 32.3 & 86.3 & 32.7 & 86.7 & 32.3 \\
Shake bottle & 100.0 & 70.0 & 100.0 & 69.0 & 100.0 & 66.3 \\
Shake bottle horizontally & 100.0 & 68.7 & 100.0 & 70.7 & 100.0 & 68.7 \\
Stack blocks three & 97.0 & 10.7 & 97.0 & 19.7 & 99.0 & 19.3 \\
Stack blocks two & 99.7 & 35.3 & 100.0 & 55.0 & 100.0 & 41.7 \\
Stack bowls three & 84.0 & 58.0 & 83.3 & 57.7 & 85.0 & 58.3 \\
Stack bowls two & 97.3 & 73.7 & 97.0 & 75.3 & 95.3 & 70.0 \\
Stamp seal & 60.0 & 5.7 & 60.3 & 9.0 & 64.3 & 5.0 \\
Turn switch & 57.7 & 16.0 & 61.7 & 8.3 & 62.0 & 7.0 \\
\midrule
Mean over 50 tasks & 85.5 & 33.9 & 86.2 & 35.4 & 86.0 & 32.1 \\
\bottomrule
\end{tabular}

%% file: tables/eval_sweeps_iid_table.tex
\begin{tabular}{llrr}
\toprule
Model & Evaluation setting & Value & Success (\%) \\
\midrule
S & Reference & Default & 94.30 \\
S & Denoising steps & 3 & 94.50 \\
\midrule
L & Reference & Default & 97.40 \\
\midrule
B & Reference & Default & 97.60 \\
\midrule
L & Execution chunk & 5 & 97.00 \\
L & Execution chunk & 8 & 96.30 \\
\midrule
B & Execution chunk & 5 & 96.40 \\
B & Execution chunk & 8 & 97.30 \\
\midrule
L & Denoising steps & 3 & 96.45 \\
L & Denoising steps & 5 & 97.05 \\
L & Denoising steps & 20 & 96.95 \\
L & Denoising steps & 50 & 96.60 \\
\midrule
B & Denoising steps & 3 & 97.30 \\
B & Denoising steps & 5 & 97.25 \\
B & Denoising steps & 20 & 96.80 \\
B & Denoising steps & 50 & 96.80 \\
\midrule
L & Noise scale $\sigma$ & 0.25 & 96.65 \\
\bottomrule
\end{tabular}

%% file: tables/eval_sweeps_ood_table.tex
\begin{tabular}{llrr}
\toprule
Model & Evaluation setting & Value & Success (\%) \\
\midrule
L & Reference & Default & 82.26 \\
\midrule
B & Reference & Default & 78.89 \\
\midrule
L & Execution chunk & 5 & 78.97 \\
L & Execution chunk & 6 & 80.19 \\
L & Execution chunk & 8 & 81.68 \\
L & Execution chunk & 16 & 81.53 \\
L & Execution chunk & 24 & 78.46 \\
L & Execution chunk & 32 & 74.92 \\
\midrule
B & Execution chunk & 5 & 75.44 \\
B & Execution chunk & 8 & 78.46 \\
\midrule
L & Denoising steps & 3 & 82.41 \\
L & Denoising steps & 5 & 82.12 \\
L & Denoising steps & 20 & 81.88 \\
L & Denoising steps & 50 & 81.66 \\
\midrule
B & Denoising steps & 3 & 79.22 \\
\midrule
L & Noise scale $\sigma$ & 0 & 82.44 \\
L & Noise scale $\sigma$ & 0.25 & 82.24 \\
L & Noise scale $\sigma$ & 0.5 & 82.50 \\
\midrule
B & Noise scale $\sigma$ & 0.25 & 79.38 \\
\bottomrule
\end{tabular}

%% file: tables/manuscript_future_pooling_table.tex
\begin{tabular}{lrr}
\hline
Future pooling & IID & OOD \\
\hline
$2\times$ head, $4\times$ wrists & 61.24 & 15.68 \\
$4\times$ head, $8\times$ wrists & 61.96 & 18.56 \\
Global spatial pooling & 61.14 & 15.04 \\
\hline
\end{tabular}

%% file: tables/manuscript_representation_diagnostics_table.tex
\begin{tabular}{lrrr}
\hline
Training objective & Clean success (\%) & Effective rank & Action readout $R^2$ \\
\hline
Action + future prediction & 66.5 & 137.70 & 0.4315 \\
Action only & 62.8 & 48.41 & 0.7937 \\
\hline
\end{tabular}